\documentclass[runningheads]{llncs}

\usepackage{eccv}

\usepackage{eccvabbrv}

\usepackage[accsupp]{axessibility}  

\usepackage{latexsym}
\usepackage[T1]{fontenc}
\usepackage[utf8]{inputenc}
\usepackage{microtype}
\usepackage{graphicx}
\usepackage{url}
\usepackage{booktabs}
\usepackage{enumitem}
\usepackage{wrapfig}
\usepackage{algorithm}
\usepackage{algpseudocode}
\usepackage[misc]{ifsym}
\usepackage{subcaption}
\usepackage{microtype}
\usepackage{amsmath}
\usepackage{colortbl}
\usepackage{xcolor} 
\definecolor{lightgray}{rgb}{0.9,0.9,0.9}
\usepackage{caption}
\usepackage{setspace}
\usepackage{url}
\usepackage{multirow}
\usepackage{tabularx}
\usepackage{blindtext}
\usepackage{pgfplots}
\pgfplotsset{compat=1.18} 
\usepackage{tikz}
\usetikzlibrary{er,positioning,bayesnet}
\usepackage{makecell}
\usepackage{tipa}
\usepackage{siunitx}
\usepackage{nicefrac}
\usepackage{listings}
\usepackage[raster,skins, most]{tcolorbox} %
\usepackage{xltabular}
\usepackage{adjustbox}
\usepackage{xurl}
\usepackage{rotating}
\usepackage[normalem]{ulem}
\usepackage{epstopdf}
\usepackage{multirow}
\usepackage{tabularx}
\usepackage{pifont}
\usepackage{amssymb}
\usepackage{fvextra} 
\DefineVerbatimEnvironment{verbatim}{Verbatim}{
  breaklines=true,
  breakanywhere=true,
  breaksymbolleft={}
}

\usepackage[table]{xcolor}
\definecolor{degradePink}{HTML}{F8D7DA}   
\definecolor{maxGainPurple}{HTML}{E7DAFF} 
\newcommand{\pmark}{\ding{109}}

\newtcolorbox{mybox}[2][]
  {colback = black!5!white, colframe = black!75!black, fonttitle = \bfseries,
    colbacktitle = black!100!black, enhanced, before upper={\fontsize{8}{11}\obeyspaces\obeylines\selectfont}, fontupper=\selectfont,
    attach boxed title to top left={yshift=-2.2mm,xshift=4mm},
    title=#2,#1}
\definecolor{purple1}{RGB}{126, 107, 196}
\newcommand{\cmark}{\ding{51}}
\newcommand{\xmark}{\ding{55}}
\newcommand{\hlword}[2]{%
  \begingroup
  \setlength{\fboxsep}{3.8pt}%
  \colorbox{#1}{\raisebox{0pt}[0.32\baselineskip][0\baselineskip]{#2}}%
  \endgroup
}

\usepackage{hyperref}

\usepackage{orcidlink}

\begin{document}

\title{GeoBrowse: A Geolocation Benchmark\\ for Agentic Tool Use with\\ Expert-Annotated Reasoning Traces} 

\titlerunning{GeoBrowse}

\author{
Xinyu Geng$^{*}$ \and
Yanjing Xiao$^{*}$ \and 
Yuyang Zhang$^{*}$ \and
Hanwen Wang \and \\
Xinyan Liu \and
Rui Min \and
Tianqing Fang \and
Yi R. Fung$^{\dagger}$
}

\begingroup
\renewcommand\thefootnote{}

\footnotetext[1]{* Equal contribution}
\footnotetext[2]{$\dagger$ Corresponding author}

\endgroup
\authorrunning{X.~Geng et al.}

\institute{
Hong Kong University of Science and Technology \\
\email{xgengad@connect.ust.hk, yrfung@ust.hk}}

\maketitle


\begin{abstract}

 Deep research agents integrate fragmented evidence through multi-step tool use. BrowseComp offers a text-only testbed for such agents, but existing multimodal benchmarks rarely require both weak visual cues composition and BrowseComp-style multi-hop verification. Geolocation is a natural testbed because answers depend on combining multiple ambiguous visual cues and validating them with open-web evidence. Thus, we introduce \textbf{GeoBrowse}, a geolocation benchmark that combines visual reasoning with knowledge-intensive multi-hop queries. \textbf{Level~1} tests extracting and composing fragmented visual cues, and \textbf{Level~2} increases query difficulty by injecting long-tail knowledge and obfuscating key entities. 
To support evaluation, we provide an agentic workflow \textbf{GATE} with five think-with-image tools and four knowledge-intensive tools, and release expert-annotated stepwise traces grounded in verifiable evidence for trajectory-level analysis.
Experiments show that GATE outperforms direct inference and open-source agents, indicating that no-tool, search-only or image-only setups are insufficient. Gains come from coherent, level-specific tool-use plans rather than more tool calls, as they more reliably reach annotated key evidence steps and make fewer errors when integrating into the final decision. The GeoBrowse bernchmark and codes are provided in \url{https://github.com/ornamentt/GeoBrowse}

\keywords{Agentic tool use \and Multimodal reasoning \and Geolocation \and Open-web information seeking}
\end{abstract}

\section{Introduction}

\begin{figure}[h]
  \centering
  \includegraphics[width=\textwidth]{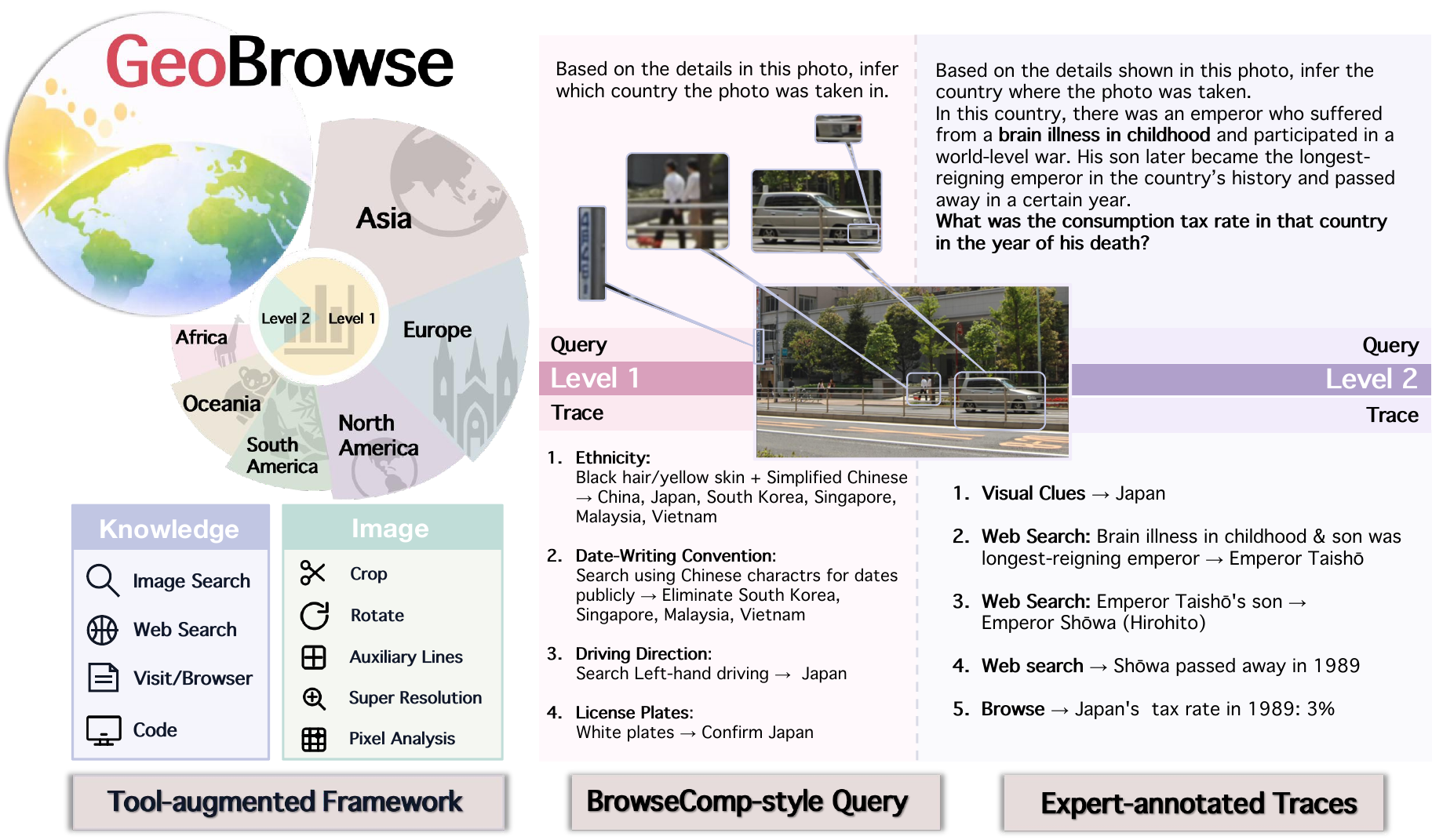}
  \vspace{-8pt}
  \caption{
  GeoBrowse couples a tool-use framework with a geolocation benchmark: Level 1 emphasizes visual cue composition, while Level 2 contains BrowseComp-style queries, all paired with expert-annotated stepwise traces.
 }
  \label{fig:main}
\end{figure}


Real-world tasks often require gathering scattered evidence from the open web and synthesizing it into a supported conclusion~\cite{nakano2021webgpt,hu2025owl}. Deep research agents tackle this by coupling Large Language Models (LLMs) with tools to plan actions, seek information, and verify evidence ~\cite{deepresearch,geminideepresearch}. To evaluate such agents, BrowseComp~\cite{wei2025browsecompsimplechallengingbenchmark} provides a text-only benchmark where evidence is intentionally fragmented and obfuscated~\cite{du2025deepresearch}, so success depends on multi-hop browsing and verification~\cite{li2025websailor,gu2025toward}. Yet, many real-world tasks are multimodal and require extracting weak visual cues before open-web evidence gathering~\cite{koh2024visualwebarena,dong2025insight}.
However, existing multimodal counterparts of BrowseComp either emphasize harder visual perception ~\cite{tao2025mmsearchplus,yuan2025visualreasoningtracerobjectlevel} or treat the image as a single step within a largely text-driven reasoning chain ~\cite{li2025benchmarkingmultimodalretrievalaugmented,geng2025webwatcher, li2025mmbrowsecomp}. 
Few benchmarks jointly stress \textbf{ambiguous visual cues} and \textbf{hard, evidence-grounded multi-hop query}, requiring agents to manipulate images to extract weak cues and then verify them through image-based retrieval, together with other tools to retrieve long-tail knowledge.

\begin{table}[t]
\centering
\caption{Comparison of existing geolocation datasets with GeoBrowse. “Local” denotes region-level coverage, whereas “Global” spans multiple continents. \textcolor{green!55!black}{\cmark} and \textcolor{red}{\xmark} indicate the presence and absence of a feature, respectively; {\pmark} indicates partial availability.}
\small
\setlength{\tabcolsep}{5pt}
\renewcommand{\arraystretch}{1.15}
\resizebox{0.9\textwidth}{!}{%
\begin{tabular}{lccccc} 
\toprule
\textbf{Dataset} &
\textbf{\shortstack[c]{Geographic\\Coverage}} &
\textbf{\shortstack[c]{Human\\Annotation}} &
\textbf{\shortstack[c]{Reasonable\\Localizability}} &
\textbf{\shortstack[c]{CoT}} &
\textbf{\shortstack[c]{Agentic\\Tool-use}} \\
\midrule
MP-16~\cite{larson2017benchmarking}              & Local  & \textcolor{red}{\xmark} & \textcolor{red}{\xmark} & \textcolor{red}{\xmark} &  \textcolor{red}{\xmark}\\
GLDv2~\cite{weyand2020google}                   & Local  & \textcolor{red}{\xmark} & \textcolor{red}{\xmark} & \textcolor{red}{\xmark} &  \textcolor{red}{\xmark}\\
VIGOR~\cite{zhu2021vigor}                        & Local & \textcolor{red}{\xmark} & \textcolor{red}{\xmark} & \textcolor{red}{\xmark} &  \textcolor{red}{\xmark}\\
Google-WS~\cite{clark2023we}                  & Global & \textcolor{red}{\xmark} & \textcolor{red}{\xmark} & \textcolor{red}{\xmark} &  \textcolor{red}{\xmark}\\
OSV~\cite{astruc2024openstreetview}          & Global & \textcolor{red}{\xmark} & \textcolor{red}{\xmark} & \textcolor{red}{\xmark} &  \textcolor{red}{\xmark}\\
GeoComp~\cite{song2025geolocation}  & Global & \textcolor{green!55!black}{\cmark} & {\pmark} & \textcolor{red}{\xmark} & \textcolor{red}{\xmark} \\
GeoVista~\cite{wang2025geovista} & Global & \textcolor{red}{\xmark} & {\pmark} & \textcolor{red}{\xmark} & {\pmark} \\
\midrule
\textbf{GeoBrowse}                               & \textbf{Global} & \textcolor{green!55!black}{\cmark} &
\textcolor{green!55!black}{\cmark} & \textcolor{green!55!black}{\cmark} & \textcolor{green!55!black}{\cmark} \\
\bottomrule
\end{tabular}%
}
\label{tab:geocomp-compare}
\end{table}

Geolocation is well suited for reasoning over ambiguous visual clues. Street-view images often contain multiple weak clues such as sign, road markings, and architectural style, where no single clue is decisive and candidates must be narrowed by combination~\cite{weyand2016planet}. Solving such cases also requires iterative reasoning and open-web verification beyond one-shot retrieval~\cite{song2025geolocation,luo2025doxing}. Together, these properties align with BrowseComp, making geolocation a natural substrate for multimodal BrowseComp-style benchmarking.

However, existing geolocation benchmarks remain limited for evaluating agentic tool use and multi-step reasoning, as summarized in Tab.~\ref{tab:geocomp-compare}. 
First, many widely used datasets such as MP-16~\cite{larson2017benchmarking}, GLDv2~\cite{weyand2020google}, VIGOR~\cite{zhu2021vigor}, Google-WS~\cite{clark2023we} and OSV~\cite{astruc2024openstreetview}, provide only coordinates or coarse labels without expert stepwise traces, which favors vision-only location prediction over tool-driven inference. 
Second, benchmarks that claim solvability such as GeoVista~\cite{wang2025geovista} and GeoComp~\cite{song2025geolocation} often filter toward salient landmarks, which removes hard cases that rely on subtle environmental and architectural signals, thus reducing overall difficulty.
Third, tool use is rarely evaluated comprehensively. Although recent GeoVista~\cite{wang2025geovista} includes zoom-in and web search, it leaves broader tools of image manipulation and browser largely untested.

We introduce \textbf{GeoBrowse}, a geolocation benchmark for evaluating agentic tool use that requires both think-with-image ability and open-web information seeking, as shown in Fig.~\ref{fig:main}. \textbf{Level~1} targets visual-first geolocation from weak and fragmented cues. Each instance includes expert stepwise annotations that list actionable visual fragments and required image manipulations. We also minimize reliance on salient landmarks to raise difficulty while remaining solvable. \textbf{Level~2} builds on the visual localization of Level~1 and further increases the difficulty of reasoning. We inject long-tail knowledge and rewrite questions into BrowseComp-style with key entities obfuscated, so solving requires multi-step retrieval and verification. Each instance is paired with an expert trace that specifies the intended evidence path.
To support comprehensive evaluation of agentic capability, we provide a \textbf{G}eolocation \textbf{A}gentic-workflow with \textbf{T}ool \textbf{E}nhancement (\textbf{GATE}) with five think-with-image tools (\textit{e.g.}, crop, rotate, pixel analysis, auxiliary lines, super-resolution) and four knowledge-oriented tools (\textit{e.g.}, web text search, web image search, visit, code interpreter).

Experiments show that GATE delivers consistent gains, outperforming direct inference across 12 Multimodal Large Language Models (MLLMs) and 3 open source agents. We find that search-only or image-only tools are insufficient, especially on Level 2 where multi-step retrieval and verification are required. Analysis of tool-use traces and ablation studies suggest that improvements come from coherent, level-specific tool-use plans rather than more tool calls: Level~1 benefits mainly from image processing, while Level~2 relies more on evidence-driven web retrieval and verification. Trajectory analysis further shows that agentic planning improves accuracy by covering more annotated key evidence and reducing final integration errors.

\paragraph{\textbf{Contributions.}} 
(1) \textbf{GeoBrowse dataset:} we present a two-level geolocation benchmark with 300 instances: Level~1 requires composing weak visual cues, and Level~2 introduces BrowseComp-style multi-hop queries with entity obfuscation. 
(2) \textbf{Expert-annotated traces:} we provide a verifiable evidence path with stepwise milestones for every instance, enabling trajectory-level analysis. 
(3) \textbf{Agentic workflow and tool suite:} we introduce \textbf{GATE}, an agentic workflow equipped with five think-with-image tools and four knowledge-intensive tools, supporting end-to-end cue extraction and open-web verification. 
(4) \textbf{Evaluation and analysis:} we benchmark 12 MLLMs and open-source agents, conduct single-tool ablations, and analyze milestone hit rates and an error taxonomy.
\begin{figure}[!t]
\noindent
\begin{minipage}[t]{0.66\linewidth}
  \centering
  \includegraphics[width=0.9\linewidth]{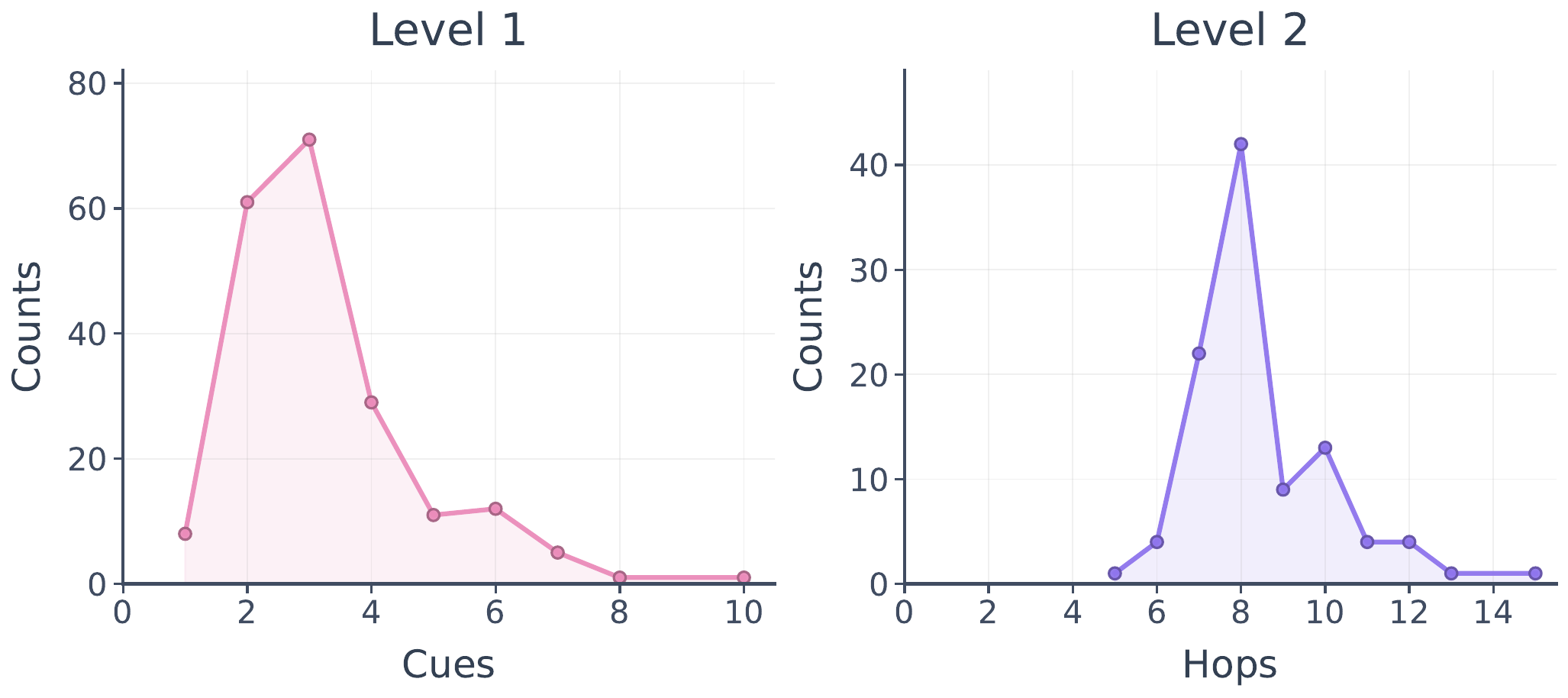}
  \vspace{-3pt}
  \captionof{figure}{Distribution of cues and hops on GeoBrowse. Cues count visual cues in Level~1 images, and hops count multi-hop steps in Level~2 queries, quantifying difficulty of visual and knowledge-intensive reasoning.}
  \label{fig:hops}
\end{minipage}%
\hfill
\begin{minipage}[t]{0.32\linewidth}
  \vspace{-100pt}
  \centering
  \captionsetup{type=table}
  \caption{Distribution of target administrative levels. The \# and \% signify count and percentage, respectively.}
  \vspace{5pt}
  \small
  \setlength{\tabcolsep}{5pt}
  \renewcommand{\arraystretch}{1.10}
  \resizebox{\linewidth}{!}{%
    \begin{tabular}{l l rr}
      \toprule
      \textbf{Target} &  & \textbf{Level 1} & \textbf{Level 2} \\
      \midrule
      \multirow{2}{*}{\textbf{Country}} & \# & 92  & 63  \\
                                      & \% & 46.2 & 62.4 \\
      \midrule
      \multirow{2}{*}{\textbf{State}}   & \# & 55  & 19  \\
                                      & \% & 27.6 & 18.8 \\
      \midrule
      \multirow{2}{*}{\textbf{City}}    & \# & 52  & 19  \\
                                      & \% & 26.2 & 18.8 \\
      \bottomrule
    \end{tabular}
  }
  \label{tab:admin_level_dist}
\end{minipage}
\end{figure}

\section{Related Work}
\paragraph{\textbf{Agentic Multimodal Tool Use.}}
Recent advances in autonomous web agents have demonstrated the potential of agentic reasoning with external tools, particularly in open-domain information seeking and synthesis~\cite{deepresearch, evolvesearch,wu2025webdancer,Li2025webthinker,song2025r1}. Extending such agents to multimodal settings further complicates reasoning, as models must integrate visual cues with textual knowledge and external verification. Prior work explores multimodal chain-of-thought prompting and structured visual reasoning~\cite{zhang2023multimodal,mitra2024compositional}, as well as grounding reasoning in external knowledge via multimodal Retrieval-Augmented Generation (RAG)~\cite{su2025thinkingimagesmultimodalreasoning, wang2025vrag, chen2025detecting}. What remains most needed is an agentic tool-use setting that tightly couples image reasoning with web search and verification. Existing benchmarks rarely test this combination: BrowseComp-style multimodal suites primarily derive difficulty from text search, with vision playing a limited role~\cite{wei2025browsecompsimplechallengingbenchmark,geng2025webwatcher,tao2025mmsearchplus}, while think-with-image benchmarks emphasize visual manipulation without comparably challenging open-web reasoning~\cite{shao2024visualcot,guo2025beyond,li2025tir,jiang2026xskill}. We bridge this gap by grounding high-difficulty, BrowseComp-like queries in geolocation, where both think-with-image operations and web search are essential for resolving fragmented visual evidence.

\paragraph{\textbf{Geolocation Benchmarks.}}
Early image-based geolocation benchmarks primarily provide coordinate supervision for recognition or retrieval, without explicitly targeting multi-step reasoning or tool use~\cite{zamir2014image,vo2017revisiting,larson2017benchmarking}. Cross-view datasets focus on matching street-level images to aerial views within constrained regions~\cite{zhu2021vigor}, while more recent large-scale street-view corpora expand geographic coverage but still emphasize single-shot localization from images~\cite{clark2023we,astruc2024openstreetview}. Recent efforts begin to incorporate human signals or limited tool interaction, such as aggregating human gameplay data~\cite{song2025geolocation} or enabling zoom and web search for agentic models~\cite{wang2025geovista}. However, existing benchmarks rarely combine globally localizable imagery with high-difficulty, information-seeking queries and expert, stepwise multimodal tool-use annotations. This leaves limited support for evaluating agentic frameworks that require both think-with-image reasoning and web-based verification under realistic constraints.

\section{GeoBrowse Benchmark}

GeoBrowse is a geolocation benchmark designed to evaluate three agentic capabilities: \emph{visual cue composition}, \emph{open-web information seeking and verification}, and \emph{tool-use planning}. It comprises 300 instances, with 199 in Level~1 and 101 in Level~2. As shown in Fig.~\ref{fig:per}, GeoBrowse spans broad geographic coverage across continents: Asia (47\%), Africa (5\%), Europe (22\%), North America (13\%), Oceania (4\%), and South America (9\%). Tab.~\ref{tab:admin_level_dist} summarizes answer granularity by administrative level. Level~2 skews toward country-level targets due to multi-hop knowledge reasoning.

\vspace{3pt}
\noindent
\begin{minipage}[t]{0.51\linewidth}
  \setlength{\parskip}{0pt}
  Level~1 targets image-centric reasoning from ambiguous cues. The number of visual cues is mostly around 2 to 4 with a mean of 3.21, whose distribution is shown in Fig.~\ref{fig:hops}. Example instances and expert-annotated traces are provided in Appendix~\ref{app:level1_case}.
  Level~2 extends Level~1 by adding multi-hop reasoning based on open-web information and cultural knowledge. Solving them perfectly requires 8.42 reasoning hops on average. Agents must first infer the location from the images, then use it as the start for iterative information seeking and verification chain. Each instance includes all Level~1 fields and an annotated evidence trace of solving multi-hop query shown in Appendix~\ref{app:level2_case}.
\end{minipage}%
\hspace{0.04\linewidth}%
\begin{minipage}[t]{0.45\linewidth}
  \vspace{-20pt}
  \centering
  \includegraphics[width=\linewidth]{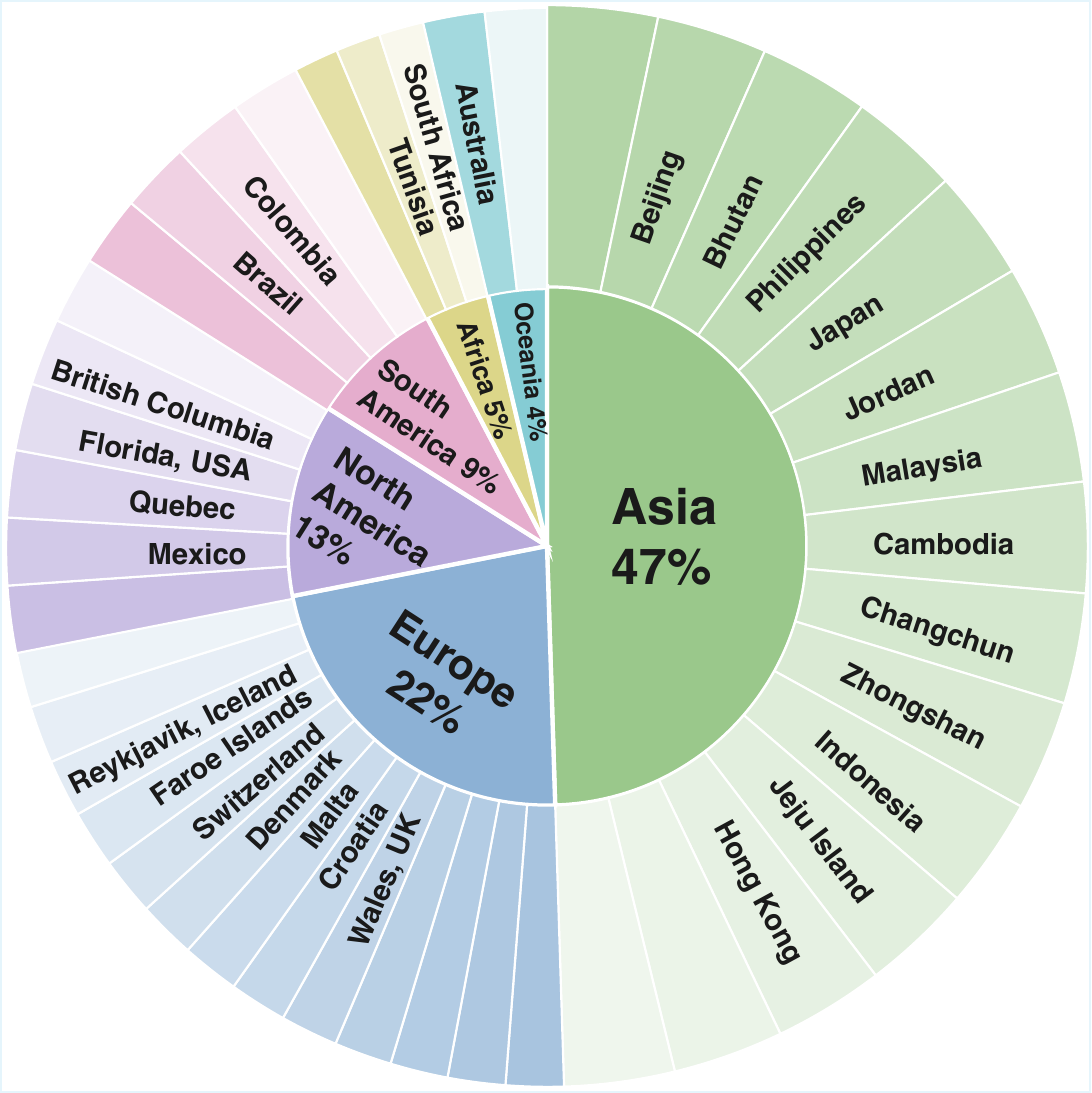}
  \captionof{figure}{Geographic coverage of GeoBrowse visual cues. The inner ring shows the percentage of instances by continent and the outer ring lists representative locations within to illustrate the diversity of covered places.}
  \label{fig:per}
\end{minipage}

\subsection{Data Collection}

\paragraph{\textbf{Level 1}}
We source raw candidates from publicly available geolocation videos created by community experts on platforms such as YouTube. These videos include stepwise reasoning, providing a natural source of geolocation inference traces. We only use videos released under permissive licenses CC BY 4.0 and retain source attribution.
From these videos, we sample representative frames for annotation. All instances are annotated by five experts with experience in geography research or geolocation. Annotators refine the candidate set by (i) identifying informative visual cues, (ii) collecting expert-level stepwise reasoning traces, and (iii) specifying appropriate tool-use strategies. These annotated candidates then undergo subsequent filtering and quality control in Sec.~\ref{sec:quality}.
To mitigate privacy concerns, all instances are filtered to avoid exposing personally identifiable information, which is detailed in Appendix~\ref{app:ethics}. 

By sourcing candidates from geolocation videos and annotating them with traces from experts, our benchmark provides content-grounded, multi-hop reasoning chains that are largely missing from prior geolocation datasets, which typically offer only coordinates or final-answer supervision.

\paragraph{\textbf{Level 2.}}

Level~2 is constructed by transforming a Level~1 geolocation query $q^{(1)}$ into a BrowseComp-style, knowledge-intensive query $q^{(2)}$. Following \cite{geng2025webwatcher,li2025websailor}, let $a^{(1)}$ denote the gold answer of the Level~1 instance. We treat $a^{(1)}$ as the root entity and build a multi-hop chain on the Wikipedia hyperlink graph $G=(\mathcal{E},\mathcal{R})$, where $\mathcal{E}$ is the set of Wikipedia entities (title of pages) and $(e,e')\in\mathcal{R}$ indicates a directed hyperlink from $e$ to $e'$. For clarity, define the outgoing-neighbor operator\begin{equation}
\mathrm{Out}(e) = \{e' \in \mathcal{E} \mid (e,e') \in \mathcal{R}\}.
\end{equation}
Starting from $e_0=a^{(1)}$, we sample a path \(\pi\)
\begin{equation}
\label{eq:graph}
\begin{aligned}
\pi &= (e_0, e_1, \ldots, e_K), \\
\text{s.t.}\quad & e_{i+1}\in \mathrm{Out}(e_i), \\
\quad &\forall i\in\{0,\ldots,K-1\}.
\end{aligned}
\end{equation}
where $K$ controls the reasoning depth. We set the terminal entity as the Level~2 answer, $a^{(2)} = e_K$. Importantly, the hops in 
$\pi$ are designed to be dependency-critical. Each step provides the minimal evidence needed to identify the next, so omitting or skipping a hop typically breaks downstream retrieval and verification, yielding an almost canonical solution path for following annotating process.

Given $\pi$, we generate the query $q^{(2)}$ through Gemini-3-Pro~\cite{gemini3} under a strict constraint: all intermediate entities \(\{e_i\}_{i=1}^{K-1}\) must be obfuscated with indirect descriptions and contextual clues rather than literal entity names, ensuring that the query cannot be solved via direct search without reasoning. The prompt of query generation is in Appendix~\ref{app:level2query}.

This construction requires an agent to first infer $a^{(1)}$ from the image to anchor the chain, and then perform multi-hop retrieval and verification to reach $a^{(2)}$. As a result, Level~2 jointly evaluates visual geolocation, knowledge-intensive reasoning, and tool-based planning.

\subsection{Quality Control and Reliability}
\label{sec:quality}


In the beginning, we use LLM with tool access to judge quality of queries. Each instance is evaluated by OpenAI o4-mini~\cite{openai_o4mini_system_card_2025} four times. We drop items solved in all runs as too easy and manually review items failed in all runs, discarding those with insufficient or incoherent evidence. This process filters both trivial and Incomplete cases and improves the reliability of the benchmark. Importantly, o4-mini is not used in any subsequent evaluation, thus avoiding model leakage.

Then, all instances are validated by three experts with experience in geography-related research and geolocation problem solving. They deem an instance reliable only if the final answer is correct and supported by the verifiable evidence trace, which is internally consistent and reproducible by our tools. For Level~2, annotators additionally check that the query preserves ambiguity by avoiding explicit named-entity leakage while still admitting a valid multi-hop reasoning path. Detailed criteria and the validation protocol are provided in Appendix~\ref{app:anno}.

Each instance is annotated by one expert and reviewed by a second expert. Disagreements are resolved by a third expert based on the completeness and verifiability of evidence. As a result, every instance receives at least two rounds of expert review. We report additional annotation details and agreement statistics in Appendix~\ref{app:anno}.

\begin{figure}[t]
  \centering
  \includegraphics[width=\textwidth]{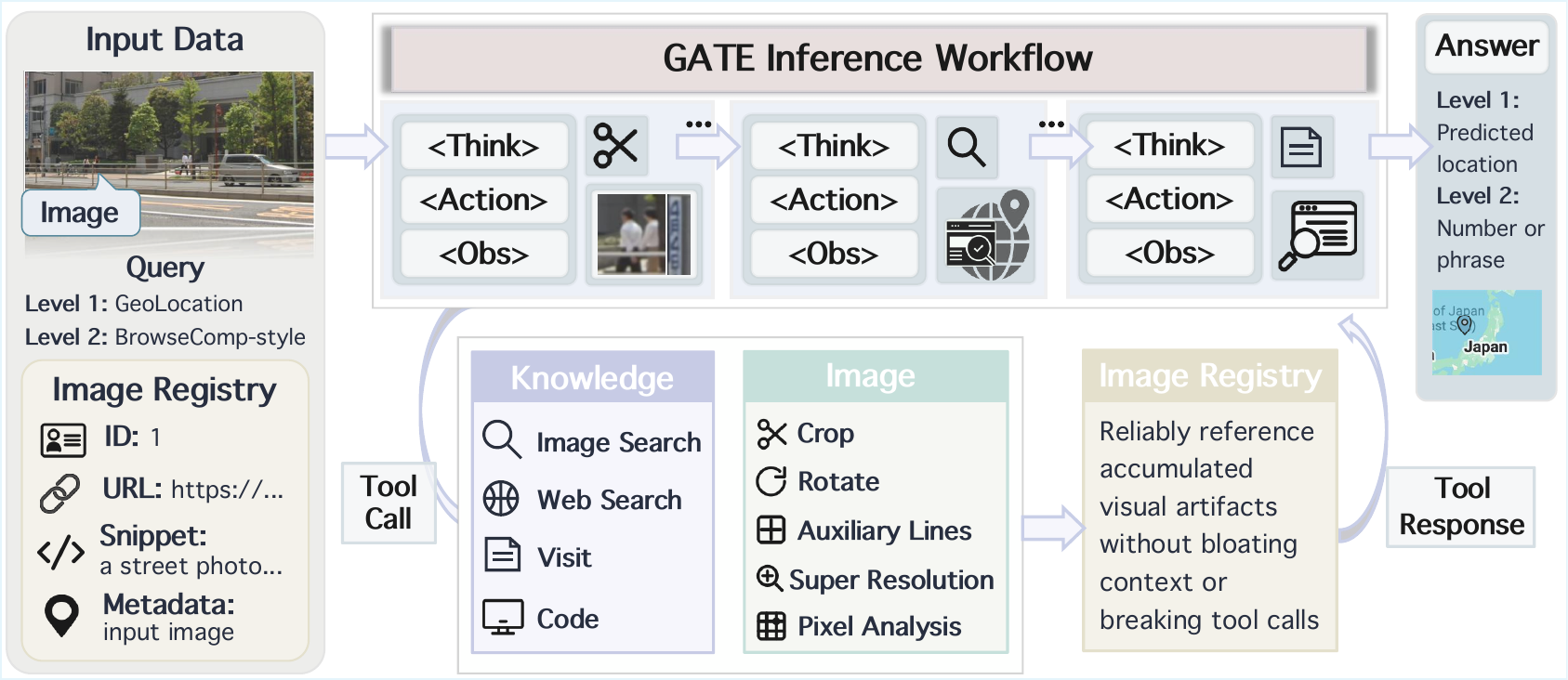}
  \caption{The pipeline of GATE, our proposed Geolocation Agentic-workflow with Tool Enhancement approach. The input image is first registered into stable \texttt{img\_id} references. GATE then follows a ReAct-style loop: \texttt{<Think>} summarizes the latest evidence and plans the next step, \texttt{<Action>} invokes an image or knowledge tool, and the tool response is returned as \texttt{<Obs>} to update the agent state. Any new images in \texttt{<Obs>} are re-registered, and the loop repeats until the final answer.}
  \label{fig:infra}
\end{figure}

\section{\textbf{G}eolocation \textbf{A}gentic-workflow with \textbf{T}ool \textbf{E}nhancement}
\vspace{-0.5em}
To address the limitations of pure LLM in grounded reasoning, such as missing fine-grained visual cues and unreliable factual verification, we augment the model with a structured tool suite and introduce our \textbf{G}eolocation \textbf{A}gentic-workflow with \textbf{T}ool \textbf{E}nhancement (\textbf{GATE}). These tools provide external perceptual and knowledge support, enabling more reliable multi-modal reasoning.
\subsection{Tools}
\label{sec:3.1}
As shown in Fig.~\ref{fig:infra}, we group our tool suite into two categories: \emph{think-with-image} tools for targeted visual cue extraction, and \emph{knowledge} tools for open-web retrieval and verification. The formal tool definitions, interfaces, and implementation details are provided in Appendix~\ref{app:tool}.

\paragraph{\textbf{Think-with-image tools.}}

These tools operate directly on the input image to expose weak and localized clues that are hard to access with a single forward pass. We support five primitive tools:
\begin{enumerate}[label=\arabic*)]
  \item \textbf{Crop} for isolating salient subregions and reducing background clutter,
  \item \textbf{Rotate} for correcting camera orientation and aligning text or structural cues, 
  \item \textbf{Auxiliary Lines} for overlaying reference guides such as vanishing lines or heading aids, 
  \item \textbf{Local Super-Resolution} for selective enhancement of small regions, 
  \item \textbf{Pixel Analysis} to figure out color statistics, edge density and texture cues. 
\end{enumerate}
Think-with-image tools are implemented in a Python sandbox~\cite{zhao2025pyvision}. When image manipulation is needed, the agent writes executable code and invokes the corresponding tool. The sandbox returns either an edited image or a brief analyzing report with statistics. For reliability and flexibility, the system prompt includes runnable reference snippets for each primitive tool. The sandbox also ships with standard libraries such as Pillow~\cite{clark2015pillow} and OpenCV~\cite{bradski2000opencv}. 

\paragraph{\textbf{Knowledge tools.}}
Knowledge tools enable evidence gathering and disambiguation, which is particularly important for BrowseComp-style Level 2 instances where multiple weak cues must be validated against external sources. We include: 
\begin{enumerate}[label=\arabic*)]
\item \textbf{Web Image Search} via Google SerpApi~\cite{serpapi} for retrieving relevant images with captions and URLs,
\item \textbf{Web Text Search} for open-domain information seeking, 
\item \textbf{Visit} via Jina~\cite{jina} for navigating specific URLs and goal-conditioned page reading,
\item \textbf{Code Interpreter}, following the implementation in~\cite{bytedanceseedfoundationcodeteam2025fullstackbenchevaluatingllms}, for symbolic computation and numerical reasoning.
\end{enumerate}

\subsection{GATE the Agentic Workflow}
\label{sec:3.2}


\paragraph{\textbf{Action space and trajectories.}}
We model query answering as an interactive ReAct-style~\cite{yao2023react} trajectory \(\tau\). As shown in Fig.~\ref{fig:infra}, at each step, the agent conditions on the accumulated history and produces a structured turn consisting of (1) a \textbf{Thought}, summarizing experience in the last step and choosing which tool to call, wrapped in \texttt{<think>...</think>}; (2) an \textbf{Action}, either a tool invocation in \texttt{<tool\_call>...</tool\_call>} or a terminal response in \texttt{<answer>...</answer>}; and (3) an \textbf{Observation}, i.e., the return of the environment, wrapped in \texttt{<tool\_response>...</tool\_response>}. 

Let \(\mathcal{A}\) denote the discrete tool-action set, and let \textbf{Finish} terminate the episode by emitting the final answer. The trajectory $\tau$ of length \(L\) is
\begin{equation}
\label{eq:traj}
\tau = \{(a_0, o_0), (a_1, o_1), \ldots, (a_L, o_L)\},
\end{equation}
where \(a_i \in \mathcal{A}\cup\{\textbf{Finish}\}\) and \(o_i\) is the observation returned after executing \(a_i\).

\paragraph{\textbf{In-trajectory image registry.}}
GATE needs to preserve images produced in a trajectory, since both image tools and web image search can generate new visual artifacts. 
However, directly appending image paths, URLs or embeddings to trajectory is impractical under bounded context window and I/O budget, and long histories can also increase cost and reduce reliability. More importantly, once images accumulate, the agent must repeatedly restate long pointers to select a specific image, where small formatting errors can make the image inaccessible and break downstream tool calls.
We maintain an in-trajectory image registry as a persistent visual state.


\vspace{2pt}
\noindent
\begin{minipage}[t]{0.61\linewidth}
  \setlength{\parskip}{0pt}
    As shown in Tab.~\ref{tab:image_regis}, each registry stores an \texttt{img\_id}, a resolvable pointer (\textit{i.e.}, URLs or Paths), a short snippet, and provenance metadata including producing tool and parent image. When a tool produces a new image, the environment registers it and exposes the updated registry to agent at the next step. The agent references images by \texttt{img\_id}, and the environment resolves an identifier to underlying pointer for subsequent inspection or tool calls. For example, Tab.~\ref{tab:image_regis} shows a simple lineage where \texttt{img\_id=0} is cropped into \texttt{img\_id=1} and then super-resolved into \texttt{img\_id=2}, with \texttt{from} explicitly tracking provenance. This avoids copying long paths, reduces hallucinated references, and improves the robustness of multimodal tool use.
\end{minipage}%
\hspace{0.03\linewidth}%
\begin{minipage}[t]{0.36\linewidth}
  \vspace{-25pt}
  \centering
  \captionof{table}{Example of image registration table.}
    \small
    \setlength{\tabcolsep}{6pt}
    \renewcommand{\arraystretch}{1.15}
    \resizebox{\linewidth}{!}{%
    \begin{tabular}{l l}
    \toprule
    \textbf{Field} & \textbf{Value} \\
    \midrule
    \textbf{img\_id}  & 0 \\
    pointer   & \texttt{input.png} \\
    snippet   & input image \\
    metadata      & input \\
    \midrule
    \textbf{img\_id}  & 1 \\
    pointer   & \texttt{crop\_1.png} \\
    snippet   & sign region crop \\
    metadata      & \texttt{Crop(img\_id=0)} \\
    \midrule
    \textbf{img\_id}  & 2 \\
    pointer   & \texttt{sr\_1.png} \\
    snippet   & 2$\times$ SR on sign \\
    metadata      & \texttt{SuperRes(img\_id=1)} \\
    \bottomrule
    \end{tabular}%
    }
    \label{tab:image_regis}
\end{minipage}

\section{Experiments}

\subsection{Experimental Setup}
\vspace{-5pt}
\textbf{\paragraph{Baselines.}}
We compare three classes of baselines on GeoBrowse:

\noindent\textbf{(1) Direct Inference.}
Models give answers in a single pass without tool calls. 
We evaluate GPT-4o~\cite{gpt4o}, GPT-4.1~\cite{gpt4.1}, GPT-5~\cite{gpt5}, GPT-5.2~\cite{gpt5_2}, Gemini-2.5-Pro/Flash~\cite{gemini2.5}, Gemini-3-Pro, Claude-4.5-Opus~\cite{anthropic_claude_opus_4_5}, Llama-3.2-90B~\cite{meta_llama_3_2_90b}, Qwen-2.5-VL family (7B/32B)~\cite{Qwen2.5-VL} and Qwen-3-VL-32B~\cite{qwen3technicalreport}.

\noindent\textbf{(2) Open-source Agents.}
We include three representative open-source agents that span complementary tool capabilities. OmniSearch~\cite{li2025benchmarkingmultimodalretrievalaugmented} is search-centric. WebWatcher~\cite{geng2025webwatcher} supports end-to-end web search, browsing, and lightweight computation for multi-step verification. PyVision~\cite{zhao2025pyvision} provides a code-driven vision sandbox for flexible image manipulation. Together, they cover search-only, browse-and-verify, and image-tool use baselines.
For fair comparison and consistency with common practice in these work, we use the default tool interfaces and GPT-4o as the backbone.

\noindent\textbf{(3) GATE.}
We evaluate GATE across backbones with standardized tool use via a fixed protocol, tool suite, and call budget. All models use the same interfaces, return formats, and image registry. The complete system prompt is in Appendix~\ref{app:sys_prompt}.
\paragraph{\textbf{Metrics and Judging.}}
We report accuracy as the primary metric. For a test set of $n$ instances, we compute
\begin{equation}
    \text{pass@1} = \frac{1}{n} \sum_{i=1}^{n} p_i,
\end{equation}
where $p_i \in \{0,1\}$ indicates whether the $i$-th prediction is correct. Because both gold answers and model outputs are short and concrete, the correctness can be determined using \textit{LLM-as-judge}~\cite{DBLP:conf/coling/LiuYHZHWDSZ24}, whose prompt follows Humanity's Last Exam~\cite{phan2025humanity} in Appendix~\ref{app:evap}.
To ensure reliability, we compare human decisions against the LLM-as-judge and report agreement statistics in Appendix~\ref{app:llmasjudge}.

\begin{table}[t]
\centering
\caption{Main results on GeoBrowse. All accuracy scores are reported as percentages. Each entry is averaged over 3 runs. \textit{Avg.} is the macro-average over city/state/country.}
\scriptsize
\resizebox{0.9\textwidth}{!}{
\begin{tabular}{l|cccc|cccc}
\toprule
\multirow{2}{*}{\textbf{Backbone}} 
& \multicolumn{4}{c|}{\textbf{Level 1: Visual Cues}} 
& \multicolumn{4}{c}{\textbf{Level 2: BrowseComp-style}} \\
\cmidrule(lr){2-5}\cmidrule(lr){6-9}
& City & State & Country & Avg. 
& City & State & Country & Avg. \\
\midrule

\rowcolor{gray!33}\multicolumn{9}{c}{\emph{\textbf{Direct Inference}}} \\
\midrule
GPT-4o              & 5.8  & 16.4 & 35.7 & 23.1 & 10.5 & 15.8 & 11.1 & 11.9 \\
GPT-4.1             & 5.8  & 14.5 & 30.4 & 19.6 & 5.3  & 5.3  & 15.9 & 11.9 \\
GPT-5               & 11.5 & 21.8 & 38.0 & 27.6 & 15.8 & 10.8 & 20.6 & 17.8 \\
GPT-5.2             & 7.7  & 9.1  & 30.4 & 18.6 & 21.1 & 21.1 & 20.6 & 20.8 \\
Gemini-2.5-Pro      & 11.5 & 21.8 & 41.3 & 28.1 & 10.8 & 10.8 & 17.5 & 14.8 \\
Gemini-2.5-flash    & 7.7  & 18.2 & 40.2 & 22.6 & 5.3  & 10.8 & 9.9  & 12.9 \\
Gemini-3-Pro        & 13.5 & 27.3 & 53.3 & 35.7 & 21.1 & 21.1 & 23.8 & 22.8 \\
Claude-4.5-Opus     & 13.5 & 23.6 & 43.4 & 30.2 & 21.1 & 26.3 & 27.0 & 25.7 \\
Llama-3.2-90B       & 3.8 & 10.9  & 19.6 & 13.1 & 5.3  & 5.3  & 9.5  & 7.9 \\
Qwen-3-VL-32B       & 7.7 & 16.4  & 16.3 & 14.1 & 0.0  & 5.3  & 9.5  & 6.9 \\
Qwen-2.5-VL-7B      & 1.9  & 6.3  & 11.1 &  7.5 & 0.0  & 0.0  & 3.2  &  1.9  \\
Qwen-2.5-VL-32B     & 3.8  & 9.1 & 17.3 & 11.1 & 0.0  & 0.0  & 6.3  &  4.0  \\
\midrule
\rowcolor{gray!33}\multicolumn{9}{c}{\emph{\textbf{Open Source Agents}}} \\
\midrule
OmniSearch  & 9.6 & 18.2  & 37.0  & 24.6  & 15.8  & 21.1 & 19.0 & 18.8 \\
WebWatcher     & 7.7 & 18.2  & 38.0  & 24.6  & 15.8  & 26.3  & 20.6  & 20.8 \\
Pyvision   & 15.4 & 20.0 & 43.5  & 29.6  & 10.5  & 15.8  & 14.3 & 13.9 \\
\midrule
\rowcolor{gray!33}\multicolumn{9}{c}{\emph{\textbf{GATE}}} \\
\midrule
GPT-4o              & 7.7  & 20.0 & 52.2 & 31.8 & 15.8 & 21.1 & 22.2 & 21.8 \\
GPT-4.1             & 7.7  & 20.0 & 45.7 & 28.6 & 15.8 & 21.1 & 23.8 & 22.8 \\
GPT-5               & 11.5 & 23.6 & 65.2 & 39.7 & 21.1 & 31.6 & 31.8 & 30.7 \\
Gemini-2.5-Pro      & 13.5 & 23.6 & 52.2 & 35.7 & 15.8 & 36.8 & 30.2 & 29.7 \\
Gemini-3-Pro        & 24.0 & 42.6 & 65.2 & \textbf{48.2} & 26.3 & 36.8 & 34.9 & \textbf{34.7} \\
Claude-4.5-Opus     & 15.4 & 32.7 & 63.0 & 42.2 & 21.1 & 36.8 & 33.3 & 32.7 \\
Llama-3.2-90B       & 3.8 & 10.9 & 19.6 & 13.1  & 5.3 & 10.5 & 12.7 & 11.9 \\
Qwen-3-VL-32B       & 7.7  & 12.7 & 30.4 & 19.6 & 5.3 & 5.3  & 9.5 & 8.9 \\
Qwen-2.5-VL-7B      & 3.8  & 9.1  & 13.0  & 9.5  & 0.0  & 5.3  & 6.4  & 4.9 \\

\bottomrule
\end{tabular}
}
\label{tab:main2}
\end{table}

\subsection{Main Results}
Tab.~\ref{tab:main2} shows that \textbf{GATE} consistently improves over direct backbone inference and also surpasses representative open-source agents.
While these agents benefit from external tools, their gains are more limited as they cover a narrower subset of tool functionalities.
On the same backbone, GATE improves GPT-4o from 23.1\% to 31.8\% on Level~1 and from 11.9\% to 21.8\% on Level~2, exceeding search-centric OmniSearch at 24.6\% and 18.8\% and vision-centric PyVision at 29.6\% and 13.9\%. This suggests search-only or image-only tools are insufficient, especially on Level~2 where multi-step retrieval and verification are required.

Breaking results down by target granularity, fine-grained localization is substantially harder: city-level accuracy of Level~1 is generally lower than state- and country-level accuracy. Besides, Level~2 city and state splits each contain 19 questions, thus a 5.2\% change corresponds to only one additional solved instance. This indicates that for harder fine-grained locations, GATE yields modest absolute improvements for most models on Level~2. 
This pattern is consistent with our Level~2 design: each instance uses geolocation as the root, and failure of localization prevents the agent from reliably initiating the subsequent multi-hop reasoning chain. The obfuscated,  BrowseComp-style clues make it difficult to back-infer the correct location.

Overall, GATE with the \textbf{Gemini-3-Pro} backbone achieves the best performance, reaching 48.2\% on Level~1 and 34.7\% on Level~2, followed by Claude-4.5-Opus and GPT-5.
GATE provides large absolute gains for strong models, improving Gemini-3-Pro from 35.7\% to 48.2\% on Level~1 and from 22.8\% to 34.7\% on Level~2.
Moreover, gains are larger on Level~1, consistent with the importance of image-side processing, while Level~2 improvements are smaller and more model-dependent, reflecting the added difficulty of long-tail knowledge retrieval and evidence integration. We therefore analyze tool traces and single-tool ablations to localize where the improvements originate.

\vspace{-3pt}
\subsection{Analysis}

\begin{figure}[t]
  \centering
  \includegraphics[width=\textwidth]{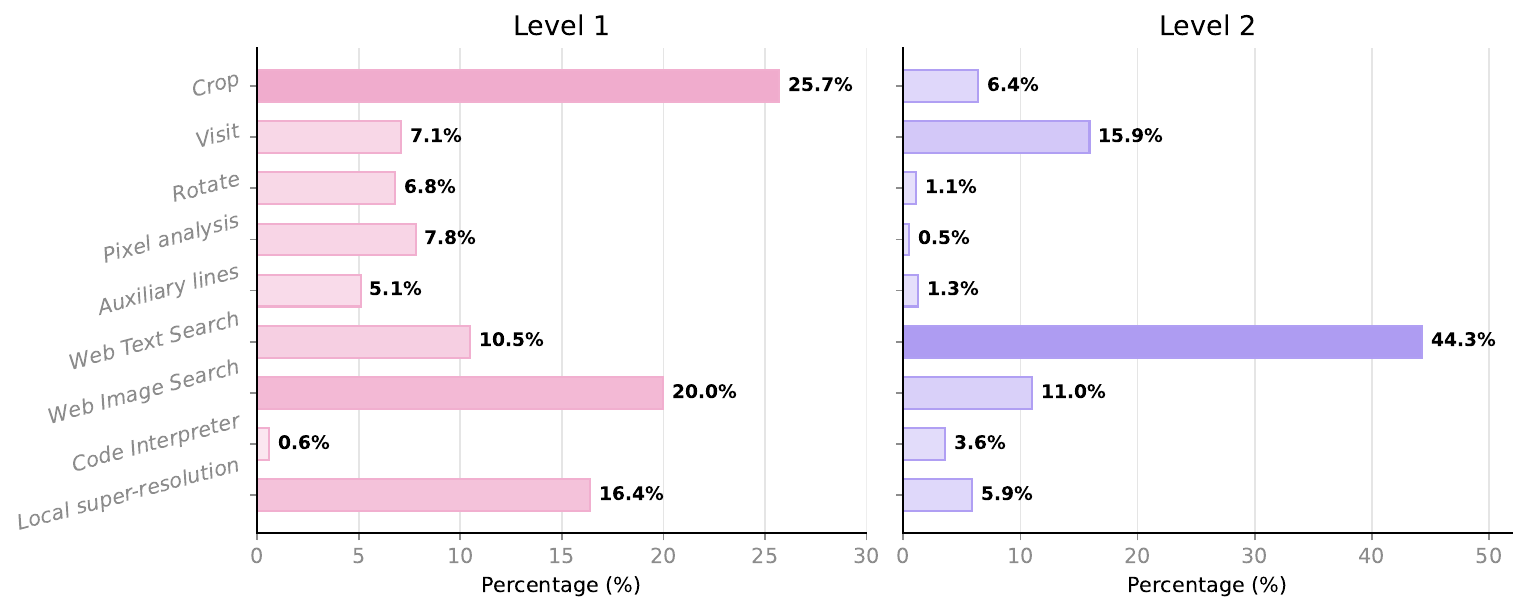}
  \caption{Tool-use distribution on GeoBrowse. Statistics are aggregated over all tool calls produced by GATE with the \textbf{Gemini-3-Pro} backbone, across Level~1 (geolocation tasks) and Level~2 (multi-step reasoning tasks requiring external knowledge).}
 
  \label{fig:tool_usage}
\end{figure}

\paragraph{\textbf{Tool Usage Profile}}
Fig.~\ref{fig:tool_usage} characterizes how agents allocate tool calls. 
Percentages denote the proportion of total tool invocations within each level. 
Level~1 concentrates on visual preprocessing and image-based retrieval, with \textit{Crop} accounting for 25.7\%, \textit{Web Image Search} 20.0\%, and \textit{Local Super-resolution} 16.4\%. 
Level~2 shifts toward web evidence gathering, where \textit{Web Text Search} accounts for 44.3\% of all tool calls and \textit{Visit} increases to 15.9\%.

Call frequency, however, does not directly measure utility. We therefore run single-tool ablations and quantify coordination with the \textbf{synergy gap}, defined as the all-tools accuracy minus the best single-tool accuracy.
Since \textit{Visit} depends on retrieved URLs, we pair it with the corresponding search tool. 
As shown in Tab.~\ref{tab:single_tool_ablation}, on Level~1, \textit{Image Processor} reaches 38.6\%, only 1.1\% below the all-tools score 39.7\%, indicating strong visual clue extraction captures most of the attainable gain. On Level~2, the best single-tool setting is \textit{Web Text Search + Visit} at 27.0\%.

\vspace{2pt}
\noindent
\begin{minipage}[t]{0.47\linewidth}
  \setlength{\parskip}{0pt}
   Level~2 leaves a larger 3.7\% gap to the all-tools score 30.7\%, suggesting a stronger dependence on tool coordination.
  This pattern matches GeoBrowse's design, where Level~1 emphasizes image-side cue extraction and Level~2 adds knowledge-intensive multi-hop retrieval and verification.
Finally, \textit{Code} alone underperforms Level~1 baseline, indicating that computation without external evidence is not effective for this task.
\vspace{4pt}
\end{minipage}%
\hspace{0.03\linewidth}%
\begin{minipage}[t]{0.50\linewidth}
  \vspace{-18pt}
    \centering
    \captionof{table}{Single-tool ablation results. Accuracy (\%) generated by GPT-5.
    \hlword{maxGainPurple}{Purple} marks the best single-tool setting per level, and
    \hlword{degradePink}{Pink} indicates degradation.}
    \vspace{3pt}
    \small
    \resizebox{\linewidth}{!}{%
    \begin{tabular}{lcc}
    \toprule
    Single-tool setting & Level 1 (\%) & Level 2 (\%) \\
    \midrule
    Baseline & 26.6 & 17.8 \\
    All tools enabled & 39.7 & 30.7 \\
    \midrule
    Web Text Search + Visit & 28.6 & \colorbox{maxGainPurple}{27.0} \\
    Web Image Search + Visit & 35.7 & 22.9 \\
    Image Processor & \colorbox{maxGainPurple}{38.6} & 21.2 \\
    Code Interpreter & \colorbox{degradePink}{23.1} & 18.6 \\
    \midrule
    \emph{Synergy gap} & 1.1 & 3.7 \\
    \bottomrule
    \end{tabular}}
    \label{tab:single_tool_ablation}
\end{minipage}
\vspace{2pt}

\paragraph{\textbf{Fixed Policy vs.\ Agentic Planning}}
\label{sec:fixed_vs_planned}




We compare the agentic planning with a fixed tool-use policy, which is constructed by fixing a representative tool chain observed in trajectories, using the same tool suite but removing all adaptive decisions and retries. Details are in Appendix~\ref{app:fix}.
Tab.~\ref{tab:fixed_vs_agentic} leads to three findings:

\noindent\textbf{(1) Fixed policy captures some of the tool gain, but remains below agentic planning.}
Fixed tool use cannot recover from early errors and is brittle to top-1 retrieval noise and ambiguous entities, lacking query refinement and tool reordering that would waste budget on plausible but irrelevant candidates.

\noindent\textbf{(2) The gap between fixed and agentic widens on Level~2.}
For example, agentic planning of GPT-4o adds 6.2\% accuracy over the fixed policy on Level~2, compared to 3.4\% on Level~1.
Level~2 requires more multi-hop retrieval and explicit verification, where adaptive planning and evidence switching provide larger returns.

\vspace{2pt}
\noindent
\begin{minipage}[t]{0.41\linewidth}
  \setlength{\parskip}{0pt}
\noindent\textbf{(3) Benefits depend on model strength, and weak models may degrade.}
While strong models consistently benefit from fixed and agentic tool use, Qwen-2.5-VL-7B drops under both regimes on Level~1 (7.5\% to 6.1\% with fixed; 7.5\% to 6.5\% with agentic), suggesting strong models filter noisy tool outputs and integrate evidence across steps, while weaker models can be misled by noisy observation.
\vspace{4pt}
\end{minipage}%
\hspace{0.03\linewidth}%
\begin{minipage}[t]{0.56\linewidth}
  \vspace{-16pt}
  \centering
  \captionof{table}{Accuracy comparison of three tool regimes.
\emph{B}: Baseline without tools. 
\emph{F}: Fixed tool-use policy.
\emph{A}: Agentic tool-use.
\hlword{maxGainPurple}{Purple} highlights strong performance, and \hlword{degradePink}{Pink} indicates performance degradation.}
\vspace{3pt}
    \small
    \setlength{\tabcolsep}{6pt}
    \renewcommand{\arraystretch}{1.15}
    \resizebox{\linewidth}{!}{%
    \begin{tabular}{lrrrrrr}
    \toprule
    & \multicolumn{3}{c}{Level 1 (\%)} & \multicolumn{3}{c}{Level 2 (\%)} \\
    \cmidrule(lr){2-4}\cmidrule(lr){5-7}
    Model & B & F & A & B & F & A \\
    \midrule
    GPT-4o          & 23.1 & 28.4 & 31.8 & 11.9 & 15.6 & 21.8 \\
    GPT-5           & 26.6 & 34.9 & 39.7 & 17.8 & 23.8 & 30.7 \\
    Gemini-3-Pro    & 28.1 & \cellcolor{maxGainPurple}40.8 & \cellcolor{maxGainPurple}48.2 & 22.8 & \cellcolor{maxGainPurple}29.2 & \cellcolor{maxGainPurple}34.7 \\
    Claude-4.5-Opus & 30.2 & 37.6 & 42.2 & 25.7 & 28.0 & 32.7 \\
    Qwen-3-VL-32B   & 14.1 & 16.8 & 19.6 &  6.9 &  6.9 &  8.9 \\
    Qwen-2.5-VL-7B  &  7.5 &  \cellcolor{degradePink}6.1 &  \cellcolor{degradePink}6.5 &  3.0 &  \cellcolor{degradePink}2.7 &  4.9 \\
    \bottomrule
    \end{tabular}
    }
    \label{tab:fixed_vs_agentic}
\end{minipage}
\vspace{2pt}

\paragraph{\textbf{Trajectory-level Analysis}}
\label{sec:cot_analysis}



Last paragraph shows that agentic planning outperforms the fixed policy. 
To test whether these gains reflect better planning rather than trial-and-error, we analyze agentic trajectories on Level~2 using human-annotated milestones.
We extract milestones from the expert traces, each specifying a key visual cue or a retrievable entity, and count a milestone hit if it appears in tool responses. Each instance has 10 milestones on average.
Tab.~\ref{tab:traj_correct_vs_incorrect_L2} reveals two findings:

\noindent\textbf{(1) Strong and weak models fail at different stages.}
For GPT-5 and Gemini-3-Pro, correct runs reach 78--80\% milestone hit rate, while incorrect runs still retain 57--59\%, indicating failures after substantial evidence is retrieved, typically during synthesis, candidate selection, or verification. 
For weaker models, milestone coverage drops sharply when incorrect (29\% for Qwen-3-VL-32B and 15\% for Qwen-2.5-VL-7B), suggesting earlier breakdowns in extracting or grounding retrievable entities.

\vspace{2pt}
\noindent
\begin{minipage}[t]{0.46\linewidth}
  \setlength{\parskip}{0pt}
\noindent\textbf{(2) Incorrect runs call more tools but do not convert them into evidence.}
For GPT-5 and Gemini-3-Pro, tool calls increase from 12.2 to 15.7 and from 12.8 to 16.1 in incorrect cases, but milestone hit rates remain far lower than in correct cases, indicating inefficient recovery after early noise.
Weaker models show smaller call increases with limited milestone coverage.
Overall, the gains from agentic tool use are better explained by more reliable milestone coverage and effective recovery.
\vspace{2pt}
\end{minipage}%
\hspace{0.04\linewidth}%
\begin{minipage}[t]{0.50\linewidth}
  \vspace{-20pt}
    \centering
    \captionof{table}{Trajectory-level analysis on Level~2 of GATE.
    \emph{Acc.} is the accuracy.
    \emph{Split} denotes correct (\emph{T}) and incorrect (\emph{F}) instances.
    \emph{MS Hit} is the milestone hit rate.}
    \small
    \resizebox{0.95\linewidth}{!}{%
    \begin{tabular}{l c l r r}
    \toprule
    Model & Acc. (\%) & Split & \makecell{MS Hit\\(\%)} & \makecell{Tool\\Calls} \\
    \midrule
    \multirow{2}{*}{GPT-5}
    & \multirow{2}{*}{30.7}
    & T   & 78.0 & 12.2 \\
    &     & F   & 57.0 & 15.7 \\
    \midrule
    \multirow{2}{*}{Gemini-3-Pro}
    & \multirow{2}{*}{34.7}
    & T   & 80.0 & 12.8 \\
    &     & F   & 59.0 & 16.1 \\
    \midrule
    \multirow{2}{*}{Qwen-3-VL-32B}
    & \multirow{2}{*}{8.9}
    & T   & 61.0 & 10.7 \\
    &     & F   & 29.0 & 12.6 \\
    \midrule
    \multirow{2}{*}{Qwen-2.5-VL-7B}
    & \multirow{2}{*}{3.0}
    & T   & 42.0 &  8.6 \\
    &     & F   & 15.0 &  9.6 \\
    \bottomrule
    \end{tabular}%
    }
    \label{tab:traj_correct_vs_incorrect_L2}
\end{minipage}

\paragraph{\textbf{Failure Diagnosis}}
To diagnose failures and inform system improvements, we label incorrect Level~2 trajectories with six coarse error types, whose detailed definitions are in Appendix~\ref{app:Failure}.

\vspace{2pt}
\noindent
\begin{minipage}[t]{0.33\linewidth}
  \setlength{\parskip}{0pt}
E1 Perception and grounding failure, E2 Retrieval strategy and querying failure, E3 Noisy or ambiguous evidence selection, E4 Missing or insufficient verification, E5 Ordering and budgeting failure, and E6 Synthesis and final decision failure.
\vspace{4pt}
\end{minipage}%
\hspace{0.03\linewidth}%
\begin{minipage}[t]{0.62\linewidth}
  \vspace{-18pt}
  \centering
  \captionof{table}{Error-type distribution (\%) on Level~2 under GATE, computed over incorrect cases only. For each model, we highlight the most frequent error.}
  \vspace{3pt}
    \small
    \setlength{\tabcolsep}{5pt}
    \renewcommand{\arraystretch}{1.15}
    \resizebox{\linewidth}{!}{%
    \begin{tabular}{lcccc}
    \toprule
    \makecell[l]{E} &
    \makecell{GPT-5} &
    \makecell{Gemini-3-Pro} &
    \makecell{Qwen3-VL-32B} &
    \makecell{Qwen2.5-VL-7B} \\
    \midrule
    E1 & 7.0  & 6.1  & 28.4 & \colorbox{degradePink}{41.4} \\
    E2 & 16.9 & 15.0 & \colorbox{degradePink}{30.0} & 32.5 \\
    E3 & 16.2 & 16.8 & 14.3 & 9.7 \\
    E4 & 13.3 & 14.9 & 7.9  & 5.2  \\
    E5 & 8.4  & 7.5  & 10.3 & 7.6  \\
    E6 & \colorbox{maxGainPurple}{38.2} & \colorbox{maxGainPurple}{39.7} & 9.1  & 3.6  \\
    \bottomrule
    \end{tabular}}
    \label{tab:l2_failure_modes_est}
\end{minipage}
\vspace{2pt}

Tab.~\ref{tab:l2_failure_modes_est} shows clear stage differences by model strength. For GPT-5 and Gemini-3-Pro, errors are dominated by \textbf{E6} at 38.2--39.7\%, indicating that failures often arise during evidence synthesis or final decision-making rather than retrieval, aligning with Finding 1 in Sec.~\ref{sec:cot_analysis}. For weaker models, most errors fall into \textbf{E1+E2} at 58.4\% for Qwen-3-VL-32B and 73.9\% for Qwen-2.5-VL-7B, reflecting difficulty in extracting retrievable entities and forming effective queries.

\section{Conclusion}
We introduced GeoBrowse, a tool-augmented geolocation benchmark designed for multi-step reasoning with fragmented visual evidence and open-web verification. GeoBrowse includes two levels: Level~1 focuses on visual-first cue composition, and Level~2 adds BrowseComp-style multi-hop knowledge reasoning. Each instance is paired with expert stepwise annotations. We also introduce an agentic workflow GATE equipped with a unified tool suite.
Our experiments show improvements come from coherent, level-specific tool-use plans rather than increased tool usage. Level~1 primarily benefits from visual processing, whereas Level~2 relies more on evidence-driven retrieval. Agentic planning further improves accuracy by hitting the annotated milestones and reducing errors in final integration.

As an expert-curated benchmark, GeoBrowse prioritizes trace quality and verifiable evidence, which currently bounds its scale and introduces some distributional skew, such as a predominance of country-level instances and uneven geographic coverage.
Future work will scale data collection via an automated pipeline that harvests candidates from permissively licensed geolocation videos, employing model-assisted pre-annotation of visual cues and draft traces, followed by selective expert review.
We will also strengthen automated quality filters by requiring tighter evidence closure, paving the way for an even more robust and comprehensive benchmark.


%
%
\newpage
\bibliographystyle{splncs04}
\bibliography{main}

\clearpage  
\appendix
\section{Ethical Considerations}
\label{app:ethics}
We follow strict ethical standards to protect privacy and prevent misuse. All visual materials are sourced from publicly available geolocation videos released under permissive licenses, with clear attribution, and we release only curated samples rather than full videos.
We apply conservative privacy screening and anonymization. Instances are filtered to avoid personally identifiable or sensitive content; identifying metadata is removed, and we do not retain IP addresses. Readable identifiers in images are blurred or masked, while only coarse, non-identifying attributes needed for geolocation are preserved.
The dataset is released solely for research. Access is governed by a data use agreement that prohibits re-identification, surveillance, harassment, or other harmful uses.

\section{Implementation Details}
\label{app:exp}

\subsection{Details of Fixed Policy}
\label{app:fix}

The fixed policy follows a deterministic pipeline: 
it generates multi-scale five-\textit{crops} at two scales and applies a single \textit{super-resolution} enhancement, performs \textit{Web Image Search} over a pre-defined list of views, then runs \textit{Web Text Search} and \textit{Visit} on URLs derived from the top image results.

In detail, given an image $I$ of size $H{\times}W$, the fixed policy applies a deterministic image-processing routine.
We generate multi-scale five-crops with scales $r\in\{0.80,\,0.60\}$, and for each $r$ we extract the standard five views, four corners and the center, with crop size $(\lfloor rH\rfloor,\lfloor rW\rfloor)$.
We rotate the image only when the EXIF orientation tag indicates a $90^\circ$ or $270^\circ$ correction; otherwise we keep the original orientation.
Each view is then enhanced using a fixed $2\times$ super-resolution operator.
We disable auxiliary overlays and pixel-level analysis in this fixed setting, since they are rarely used in practice and are less reliable to control, consistent with the low call frequency shown in Figure~\ref{fig:tool_usage}.
To keep the procedure controlled, the fixed setting disables auxiliary overlays and pixel-level analysis, and it does not allow additional calls beyond the prescribed sequence.

We run Web Image Search on the top-$K$ views in a fixed order (original $\rightarrow$ five-crops at $r{=}0.80$ $\rightarrow$ center crop at $r{=}0.60$; $K{=}7$).
For each image query, we take the top-1 result URL.
For the first $J$ unique URLs ($J{=}3$), we perform a Web Text Search using the image-result title as the query (top-1), then \texttt{Visit} that page (once).
No additional tool calls are allowed.
For Level~2, we add one extra search on the textual query, while keeping the rest of the pipeline unchanged.

\subsection{Details of Failure Diagnosis}
\label{app:Failure}
We define six common error types to categorize failure modes in agentic tool use. Each type is described by its characteristic cause and observable trajectory signals.

\begin{itemize}
  \item \textbf{E1 Perception and grounding failure.}
  The agent fails to extract key visual cues or to ground them into retrievable entities that are present in the image, such as road-sign language, partial place names, landmark shapes, flags or emblems, and distinctive road infrastructure. Typical signals include a low milestone hit rate (below 30\%) or hits dominated by vague descriptions (e.g., ``looks like Europe'') rather than concrete, retrievable entities.

  \item \textbf{E2 Retrieval strategy and querying failure.}
  Some cues are extracted, but queries are overly generic or excessively long, contain spelling errors, or emphasize the wrong keywords. The agent may also misuse tools, such as relying on image search when text search is more appropriate, or repeatedly issuing similar searches without refining the query. Typical signals are many tool calls but few strongly related entities appearing in tool responses, with a low-to-moderate milestone hit rate (20--45\%).

  \item \textbf{E3 Noisy or ambiguous evidence selection.}
  The agent retrieves multiple plausible candidates but fails to handle ambiguity, such as homonyms, near-duplicate place names, or visually similar landmarks, and ultimately commits to the wrong branch. Typical signals include a moderate or even high milestone hit rate (40--60\%), but the hit milestones align with an incorrect branch.

  \item \textbf{E4 Missing or insufficient verification.}
  The agent does not perform visits for confirmation, or it relies on snippets and a single weak source to make a decision, leaving the evidence chain incomplete. Typical signals include a moderate milestone hit rate (40--60\%) combined with the absence of decisive verification milestones, such as authoritative pages, reliable entries, or explicit matching statements.

  \item \textbf{E5 Ordering and budgeting failure.}
  The agent spends budget on unproductive steps due to poor tool ordering, such as visiting before searching, committing too early to a hypothesis, or pursuing irrelevant directions, leaving insufficient budget for critical checks. Typical signals include near-cap tool usage with stagnant or declining milestone coverage, or early stopping that misses key steps.

  \item \textbf{E6 Synthesis and final decision failure.}
  The agent hits many correct milestones, especially those where retrievable or strongly related entities appear, but fails to integrate evidence, resolve conflicts, or map the evidence to the correct country or region. Typical signals include a high milestone hit rate (at least 50\%) and elevated tool calls in incorrect cases, reflecting repeated verification attempts or oscillation in conclusions.
\end{itemize}

\begin{table}[t]
\centering
\caption{Inter-annotator agreement (IAA) on a blinded double-annotated subset.
Agree(\%) is strict city-level exact match.
$\kappa^{w}$ denotes weighted Cohen's kappa (quadratic weights) for ordinal 1--4 trace-quality ratings (verifiability/completeness/coherence averaged).}
\small
\setlength{\tabcolsep}{6pt}
\resizebox{0.8\linewidth}{!}{%
\begin{tabular}{l c c c c c c}
\toprule
Subset & N & Agree(\%) & $\kappa_{\text{country}}$ & $\kappa_{\text{state}}$ & $\kappa_{\text{city}}$ & $\kappa^{w}_{\text{trace}}$ \\
\midrule
Level 1 & 199 & 82.0 & 0.93 & 0.86 & 0.72 & 0.62 \\
Level 2 & 101 & 76.0 & 0.91 & 0.82 & 0.66 & 0.55 \\
\midrule
Macro avg. (L1/L2) & 300 & 79.0 & 0.92 & 0.84 & 0.69 & 0.59 \\
\bottomrule
\end{tabular}%
}
\label{tab:iaa}
\end{table}

\begin{table}[t]
\centering
\caption{Inter-rater agreement between LLM judges and human experts on GeoBrowse (binary 0--1 labels, $N{=}300$).
95\% Wilson confidence intervals are reported for raw agreement.}
\small
\resizebox{0.75\linewidth}{!}{%
\begin{tabular}{lccc}
\toprule
\textbf{Judge} & \textbf{Agreement (\%)} & \textbf{95\% CI} & \textbf{Cohen's $\kappa$} \\
\midrule
GPT-4o               & 95.0 & [91.9, 97.9] & 0.90 \\
Claude-3.5-Sonnet    & 91.1 & [83.9, 95.2] & 0.82 \\
Gemini-2.5-Pro       & 94.8 & [90.9, 96.5] & 0.89 \\
Qwen2.5-72B-Instruct & 93.1 & [86.4, 96.6] & 0.86 \\
\bottomrule
\end{tabular}%
}
\label{tab:human}
\end{table}

For each incorrect example, we assign a single error label using the following ordered decision procedure.

\begin{enumerate}
  \item \textbf{E1 (Perception and grounding failure).}
  If the key entities or phrases required by the \emph{correct branch} never appear in any tool response, and the milestone hit rate is very low, label the case as E1.

  \item \textbf{E2 (Retrieval strategy and querying failure).}
  If some relevant cues appear, but the queries are clearly inappropriate or the wrong tools are selected such that retrieval consistently drifts away from the target evidence, label the case as E2.

  \item \textbf{E3 (Noisy or ambiguous evidence selection).}
  If multiple candidates are retrieved and the final choice is inconsistent with the stronger evidence among them, label the case as E3.

  \item \textbf{E4 (Missing or insufficient verification).}
  If the evidence chain is clearly incomplete, for example due to missing visits or lack of cross-source verification, label the case as E4.

  \item \textbf{E5 (Ordering and budgeting failure).}
  If the budget is spent on irrelevant directions, or the tool order is clearly unreasonable such that critical verification steps are missed, label the case as E5.

  \item \textbf{E6 (Synthesis and final decision failure).}
  Otherwise, if many milestones are hit but the final answer is still incorrect or conflicts are not resolved, label the case as E6.
\end{enumerate}

\subsection{Reliability of LLM-as-Judge}
\label{app:llmasjudge}
We evaluate answers using an LLM-as-judge protocol, motivated by the fact that both the gold labels and model predictions are intentionally short (e.g., country, city, or a small categorical label). In this setting, LLM-based judging is (i) scalable and cost-effective compared to fully manual grading, (ii) more robust than exact-string matching to benign surface variation (aliases, abbreviations, alternative spellings), and (iii) capable of applying consistent normalization rules (e.g., mapping ``NYC'' to ``New York City'') while still enforcing correctness constraints.

LLM-as-Judges may be biased when used as unconstrained graders.
We therefore conduct a blinded human audit on GeoBrowse with binary 0--1 correctness in Tab.~\ref{tab:human}.
GPT-4o agrees with expert labels in 95.0\% of cases and 95\% Wilson CI [91.9, 97.9] with Cohen's $\kappa{=}0.90$, indicating strong judge--expert consistency.
Other judges show similar reliability with 91.1--94.8\% agreement and $\kappa{=}0.82$--0.89, supporting the use of constrained LLM judgments for VQA answer correctness.

\subsection{Details of Annotation}
\label{app:anno}
\paragraph{Annotator calibration.}
Before full-scale annotation, all annotators complete a calibration phase consisting of (i) reading detailed guidelines, (ii) studying worked examples with gold rationales, and (iii) performing trial annotations on a shared subset. Trial annotations are reviewed by the project leads, and only annotators who meet a predefined consistency standard proceed to full annotation.

\paragraph{Annotation guidelines and evidence requirements.}
For each instance, annotators must (i) confirm the gold answer is uniquely supported by the available evidence, (ii) write a stepwise trace that lists the key visual fragments and corresponding verification steps, and (iii) record the supporting sources used for verification. Traces are required to be verifiable, with each step grounded either in visible image evidence or in externally retrieved evidence. For Level~2, annotators additionally verify the abstraction constraint: entity names on the sampled chain must not appear verbatim in the query, while the query still preserves a valid multi-hop path that can be reconstructed through retrieval and verification.

\paragraph{Review and adjudication protocol.}
Annotation follows a multi-stage review workflow. Each instance is annotated by one expert and independently reviewed by a second expert. Any disagreement in the final answer, trace validity, or Level~2 abstraction compliance is escalated to a third expert for adjudication. Adjudication decisions are made based on the completeness and verifiability of evidence, rather than majority vote. Consequently, every instance is reviewed by at least two experts.

\paragraph{Inter-annotator agreement.}
To quantify annotation reliability beyond this review process, we additionally construct a blinded IAA test: we ask two annotators to independently provide (i) the final answer at country/state/city level, (ii) for Level~2, binary checks for entity leakage / ambiguity / valid multi-hop path, and (iii) 1–4 ordinal ratings for trace verifiability, completeness, and coherence. We report raw agreement and Cohen’s $\kappa$ in Tab.\ref{tab:iaa}.

Tab.~\ref{tab:iaa} shows strong reliability at coarse granularity, with near-perfect agreement at the country level ($\kappa=0.92$) and substantial agreement at the state level ($\kappa=0.84$). City-level consistency is lower (strict match 79.0\%; $\kappa=0.69$), reflecting the increased difficulty at finer geographic resolution. Trace-quality ratings achieve moderate agreement ($\kappa^{w}_{\text{trace}}=0.59$), indicating overall alignment with some subjectivity in qualitative judgment.

\clearpage
\onecolumn
\section{Prompts}
\label{app:prompt}
\subsection{Create Level 2 query}
\label{app:level2query}

\begin{tcolorbox}[breakable,title=\texttt{Prompt: Query Generation from a Knowledge-Graph Path}]
\textbf{Task:} Given a \emph{knowledge-graph node chain} (a single path), generate a single natural-language \textbf{query} whose reasoning follows the chain hop by hop. The \textbf{root node} must be resolved directly from the image (location identification). Each subsequent node corresponds to \textbf{one hop} needed by the query. Your output must also include (i) the \textbf{concrete entity} for \emph{every} node and (ii) the \textbf{reasoning} that links each node to the next. Finally, output the \textbf{gold answer}, which is exactly the \textbf{last node}.

\bigskip

\textbf{Notes:}
\begin{itemize}
  \item \textbf{(1) Obfuscate all intermediate nodes.} The query must not directly reveal intermediate entities (non-root, non-answer). Use vague but sufficient clues so that \emph{each hop is necessary}. Missing any intermediate hop should make the final answer unattainable.
  \item \textbf{(2) The answer must be unique and short.} It can be a number or a word, but must be uniquely determined and concise.
  \item \textbf{(3) Do not add extra clues for the root node.} The root should be obtained from the image without additional textual hints.
\end{itemize}

\bigskip

\textbf{Examples:}

\medskip

\textbf{Input:}
\begin{itemize}
  \item kg\_path:
  \begin{enumerate}
    \item Root (from image): \texttt{Ireland}
    \item Capital of the root: \texttt{Dublin}
    \item 16th-century-founded college in the capital: \texttt{Trinity College Dublin (1592)}
    \item Alumni physicist: \texttt{Ernest T.\ S.\ Walton}
    \item British collaborator: \texttt{John Cockcroft}
    \item Device named after both: \texttt{Cockcroft--Walton accelerator / voltage multiplier}
    \item Landmark experiment target: \texttt{Lithium}
    \item Produced particle: \texttt{Helium (alpha particle / helium nucleus)}
    \item First observer of the same particle via solar spectroscopy: \texttt{Pierre Janssen}
    \item Second observer (English astronomer): \texttt{Joseph Norman Lockyer}
    \item Gold answer (year elected Fellow): \texttt{1869}
  \end{enumerate}
\end{itemize}

\textbf{Output:}
\begin{verbatim}
{
  "query": "Based on the image, identify the country. In the capital of this country, there is a college founded in the 16th century. An alumnus of this college is a physicist who shared a top physics prize with a British collaborator for a landmark experiment using a device named after both of them. In that experiment, they used accelerated particles to bombard a specific element and produced the nucleus of an element that was first identified by a scientist through observing a solar 
  spectral phenomenon; in the same year, an English astronomer also observed the same phenomenon. In what year was this English astronomer elected a fellow of the Royal Society?",
  "nodes": [
    {
      "hop": 1,
      "role": "root_from_image",
      "entity": "Ireland",
      "reasoning": "The image provides geographic cues that allow the
      country to be identified as Ireland."
    },
    {
      "hop": 2,
      "role": "capital",
      "entity": "Dublin",
      "reasoning": "Given the country (Ireland), its capital city is Dublin."
    },
    {
      "hop": 3,
      "role": "college",
      "entity": "Trinity College Dublin (founded 1592)",
      "reasoning": "In Dublin, a well-known college founded in the late 16th century is Trinity College Dublin (1592)."
    },
    {
      "hop": 4,
      "role": "alumnus_scientist",
      "entity": "Ernest T. S. Walton",
      "reasoning": "A notable physicist alumnus of Trinity College Dublin is Ernest T. S. Walton."
    },
    {
      "hop": 5,
      "role": "collaborator",
      "entity": "John Cockcroft",
      "reasoning": "Walton shared a major prize with his British collaborator John Cockcroft."
    },
    {
      "hop": 6,
      "role": "device",
      "entity": "Cockcroft–Walton accelerator / voltage multiplier",
      "reasoning": "Their work is associated with the device named after both scientists: the Cockcroft–Walton accelerator (a high-voltage multiplier)."
    },
    {
      "hop": 7,
      "role": "target_element",
      "entity": "Lithium",
      "reasoning": "In the landmark experiment, accelerated particles were used to bombard lithium."
    },
    {
      "hop": 8,
      "role": "produced_particle",
      "entity": "Helium (alpha particle / helium nucleus)",
      "reasoning": "Bombarding lithium produced alpha particles, i.e., helium nuclei (helium)."
    },
    {
      "hop": 9,
      "role": "first_observer",
      "entity": "Pierre Janssen",
      "reasoning": "Helium was first identified from a solar spectral observation associated with Pierre Janssen."
    },
    {
      "hop": 10,
      "role": "second_observer",
      "entity": "Joseph Norman Lockyer",
      "reasoning": "In the same year, the English astronomer Joseph Norman Lockyer also observed the same solar spectral feature."
    },
    {
      "hop": 11,
      "role": "gold_answer",
      "entity": "1869",
      "reasoning": "Lockyer's election as a Fellow of the Royal Society occurred in 1869."
    }
  ],
  "gold_answer": "1869"
}
\end{verbatim}

\bigskip

\textbf{Here is the input you need to process:}

\begin{itemize}
  \item kg\_path: \{kg\_path\}
\end{itemize}

\textbf{Output format:}
\begin{verbatim}
{
  "query": "",
  "nodes": [
    {"hop": 1, "role": "", "entity": "", "reasoning": ""},
    {"hop": 2, "role": "", "entity": "", "reasoning": ""}
  ],
  "gold_answer": ""
}
\end{verbatim}
\end{tcolorbox}

\subsection{Quality Control}

\begin{tcolorbox}[breakable,title=\texttt{Prompt: Tool Call Rationality Evaluation}]
\textbf{Role:} You are a professional AI interaction quality assessor. Your core task is to analyze dialogue snippets between a user and an AI assistant that include a \texttt{<tool\_call>} tag followed by a \texttt{<think>} tag.

\bigskip

\textbf{Task:} Judge whether the tool call (\texttt{<tool\_call>}) is \emph{reasonable} according to the three criteria defined below. “Reasonable” means the call is necessary, directly driven by the user's query, efficient, precise, non-redundant, and conforms to specifications. Also evaluate the thought process (\texttt{<think>}) for logical accuracy and to ensure no guessing or fabrication.

\bigskip

\textbf{Evaluation Criteria:}
\begin{enumerate}
  \item \textbf{Information Non-Redundancy:}  
    The requested information or action in the tool call is \emph{not} already provided or easily derivable from prior dialogue, the user’s current question, or the assistant’s previous answers.  
    \emph{Check:} Is there any overlap or repeated request?
  \item \textbf{Goal Alignment:}  
    The tool call’s purpose and expected result directly serve the user’s explicit intent or core need in this turn.  
    \emph{Check:} Does it advance the user’s main objective?
  \item \textbf{Logical Reasoning and Accuracy:}  
    The assistant's thought process shows clear, correct logic and reliable grounding - no unfounded guesses or fabrications. The \texttt{<think>} section should be concise.  
    \emph{Check:} Is the reasoning well-structured and evidence-based?
\end{enumerate}

\bigskip

\textbf{Instruction:}  
Compare the user’s question and the model’s generated snippet (including \texttt{<tool\_call>} and \texttt{<think>}). If \emph{all} criteria are met, output:
\begin{verbatim}
A
\end{verbatim}
Otherwise (any criterion unmet or room for improvement), output:
\begin{verbatim}
B
\end{verbatim}

\bigskip

\textbf{User Question:}  
\{query\}

\medskip

\textbf{Model Generation:}  
\{model\_gen\}
\end{tcolorbox}

\subsection{System Prompt}
\label{app:sys_prompt}

\begin{tcolorbox}[breakable,title=\texttt{Prompt: Web-augmented Visual Question Answering}]
\textbf{Task:} Answer the given visual question by combining careful image understanding with thorough web information seeking. Your goal is to return an accurate, well-supported final answer.

\bigskip

\textbf{Inputs:}
\begin{itemize}
  \item \texttt{question}:~\{question\}
  \item \texttt{image}:~\{image\}
\end{itemize}

\bigskip

\textbf{Core Principles:}
\begin{itemize}
  \item \textbf{Stepwise decomposition:} Break the question into sub-questions and address them in a logical order. After each step, briefly summarize what has been established and state the next sub-question to pursue.
  \item \textbf{Detailed image description:} Describe the image carefully even if the answer seems obvious. If the input contains multiple sub-images, describe each sub-image separately.
  \item \textbf{Evidence-driven web search:} Use web search to resolve unknown entities, verify claims, and disambiguate similar candidates. Prefer reliable sources and cross-check when results are ambiguous.
  \item \textbf{Budgeted completion:} Produce a final answer within \textbf{10 turns}. If evidence remains incomplete, state the most likely answer and briefly note the remaining uncertainty.
\end{itemize}

\bigskip

\textbf{Output Requirements:}
\begin{itemize}
  \item Provide the final answer clearly and concisely at the end.
  \item When relevant, include key supporting evidence (entities, titles, or short factual statements) that justified the decision.
\end{itemize}

\bigskip

\textbf{Template:}
\begin{verbatim}
Image description:
- <detailed description; if multiple panels, describe each separately>

Progress so far:
- <1-3 bullet summary of confirmed facts>

Next sub-question:
- <the next question to resolve>

Action:
- <what you will search/verify next and why>

Final answer:
- <final answer, concise; if uncertain, state uncertainty briefly>
\end{verbatim}
\end{tcolorbox}

\subsection{Evaluation Prompt}
\label{app:evap}

\begin{tcolorbox}[breakable,title=\texttt{Prompt: Response Accuracy Evaluation}]
\textbf{Task:} Judge whether the given \texttt{response} correctly answers the \texttt{question} based on the precise and unambiguous \texttt{correct\_answer}.  

\bigskip

\textbf{Inputs:}
\begin{itemize}
  \item \texttt{question}:~\{question\}
  \item \texttt{response}:~\{response\}
  \item \texttt{correct\_answer}:~\{correct\_answer\}
\end{itemize}

\bigskip

\textbf{Output Fields:}
\begin{itemize}
  \item \textbf{extracted\_final\_answer}:  
    The exact answer string extracted from \texttt{response}.  
    If no clear final answer is present, use \texttt{None}.
  \item \textbf{reasoning}:  
    A brief explanation of why \texttt{extracted\_final\_answer} does or does not match \texttt{correct\_answer}.  
    Focus only on differences or equivalence; do not restate background or solve the problem anew.
  \item \textbf{correct}:  
    \texttt{yes} if \texttt{extracted\_final\_answer} matches \texttt{correct\_answer} (allowing small numerical margins), else \texttt{no}.
  \item \textbf{confidence}:  
    The confidence score (0\%-100\%) as given in \texttt{response}.  
    If none is provided, use \texttt{100\%}.
\end{itemize}

\bigskip

\textbf{Template:}
\begin{verbatim}
extracted_final_answer: <answer or None>
reasoning: <your brief comparison>
correct: <yes or no>
confidence: <0% to 100%>
\end{verbatim}
\end{tcolorbox}

\subsection{Tool Definition}
\label{app:tool}

\begin{tcolorbox}[title=Tool: \texttt{Code Interpreter}]
\textbf{Description:} Executes Python code for calculation, data analysis, or content extraction.

\textbf{Arguments:}
\begin{itemize}
    \item \texttt{code} (string): The Python code to execute. \textit{(Required)}
\end{itemize}
\end{tcolorbox}

\begin{tcolorbox}[title=Tool: \texttt{Web Text Search}]
\textbf{Description:} Retrieves the top 10 text excerpts from Google's text search engine using one or more search queries.

\textbf{Arguments:}
\begin{itemize}
    \item \texttt{queries} (array of strings): List of search queries. \textit{(Required)}
\end{itemize}
\end{tcolorbox}

\begin{tcolorbox}[title=Tool: \texttt{ Web Image Search}]
\textbf{Description:} Retrieves top 5 images and descriptions from Google's image search using a given image URL. Should only be used once.

\textbf{Arguments:}
\begin{itemize}
    \item \texttt{image\_urls} (array of strings): List of image URLs to search with. \textit{(Required)}
\end{itemize}
\end{tcolorbox}

\begin{tcolorbox}[title=Tool: \texttt{Visit}]
\textbf{Description:} Visits a given webpage and returns a summary based on a specified goal.

\textbf{Arguments:}
\begin{itemize}
    \item \texttt{url} (string): The target webpage URL. \textit{(Required)} 
    \item \texttt{goal} (string): The goal or information the agent seeks from the webpage. \textit{(Required)}
\end{itemize}
\end{tcolorbox}

\begin{tcolorbox}[title=Tool: \texttt{image\_processor}]
\textbf{Description:} Downloads an image from a URL and executes user-supplied code in a sandbox (Pillow/Numpy preloaded) to transform or analyze the image. Use this tool for image-producing operations; return outputs via \texttt{save\_image(img)}.

\textbf{Arguments:}
\begin{itemize}
    \item \texttt{code} (string): Python code to run on the image (with helpers preloaded). \textit{(Required)}
    \item \texttt{image\_id} (string): The ID of the image to process. \textit{(Required)}
\end{itemize}

\textbf{Core Capabilities:}
\begin{itemize}
    \item \textbf{Crop:} extract regions of interest (e.g., center crop or bbox crop).
    \item \textbf{Rotate:} correct orientation (e.g., rotate $90^\circ$ with \texttt{expand=True}).
    \item \textbf{Enhance:} improve visibility (e.g., auto-contrast, contrast scaling).
    \item \textbf{Mark:} draw shapes/text overlays for visual inspection (e.g., boxes, labels).
    \item \textbf{Analyze:} compute statistics and return structured results (e.g., mean RGB as JSON).
\end{itemize}
\end{tcolorbox}

\subsection{ReAct Trajectories}

\begin{tcolorbox}[breakable,title=Case Trajectory]
<think> thinking process here </think>\\
<tool\_call>\\
{"name": "tool name here", "arguments": {"parameter name here": parameter value here, "another parameter name here": another parameter value here, ...}}\\
</tool\_call>\\
<tool\_response>\\
tool\_response here\\
</tool\_response>\\
(more thinking processes, tool calls and tool responses here)\\
<think> thinking process here </think>\\
<answer> answer here </answer>
\end{tcolorbox}

\section{Case Study}

\subsection{Level 1 Case}
\label{app:level1_case}

\begin{tcolorbox}[breakable,title={Case Study}]

\textbf{JSONL fields.}

\begin{verbatim}
{
  "question_id": 21,
  "prompt": "Identify the dam in which Chinese city through the image.",
  "image": ".image_path/24.png",
  "gold_image_answer": "Xiangjiang Songjiazhou Hydropower Station Dam in Yongzhou City, Hunan Province",
  "image_solution": "Seen as follow",
  "source": "https://example.com"
}
\end{verbatim}

\medskip
\textbf{Image Solution.}

\begin{verbatim}
{
  "query": "Identify the dam in which Chinese city through the image.",
  "nodes": [
    {
      "hop": 1,
      "role": "root_from_image",
      "entity": "Flat, foggy river-valley setting with morning illumination cues",
      "reasoning": "The scene shows flat terrain and thin fog consistent with early-morning conditions, providing initial environmental context for downstream reasoning."
    },
    {
      "hop": 2,
      "role": "sun_direction_view_geometry",
      "entity": "Left side faces east; view direction is toward the south",
      "reasoning": "Using sunlight direction and balcony arrangement, the left side can be inferred as east, which implies the camera/view axis points south."
    },
    {
      "hop": 3,
      "role": "hydraulic_gradient",
      "entity": "Higher far-side water level implies the far (southern) side is upstream",
      "reasoning": "The dam separates two water levels; the distant side appears slightly higher, indicating it is upstream relative to the near side."
    },
    {
      "hop": 4,
      "role": "river_flow_direction",
      "entity": "River flows from south to north",
      "reasoning": "Given the upstream side is to the south and downstream is to the north, the river flow direction is south-to-north."
    },
    {
      "hop": 5,
      "role": "river_scale_estimate",
      "entity": "Main channel width roughly 300--500 meters",
      "reasoning": "Estimating from the dam gate span and count, the mainstream river width falls in the 300--500 m range, constraining candidate rivers."
    },
    {
      "hop": 6,
      "role": "navigation_inference",
      "entity": "No ship locks; low navigational capacity on this reach",
      "reasoning": "The dam lacks navigation locks, suggesting this segment is not a major navigational route, helping rule out large trunk waterways."
    },
    {
      "hop": 7,
      "role": "hydropower_indicator",
      "entity": "Left-bank power station indicates stable flow with limited annual variation",
      "reasoning": "A power plant adjacent to the dam implies dependable discharge suitable for hydropower, pointing to a river with relatively stable seasonal flow."
    },
    {
      "hop": 8,
      "role": "regional_climate_judgment",
      "entity": "Southern China",
      "reasoning": "The combination of hydropower suitability, foggy lowland morning conditions, and implied hydrology is more consistent with southern Chinese river systems."
    },
    {
      "hop": 9,
      "role": "landuse_context",
      "entity": "Industrial/commercial facilities: cement plant and grain silos on the left bank",
      "reasoning": "Visible cement-related infrastructure and grain silos indicate an industrial riverbank economy typical of regional-level cities."
    },
    {
      "hop": 10,
      "role": "urban_morphology",
      "entity": "Clusters of high-rise residential buildings suggest a regional-level city",
      "reasoning": "High-rise residential clusters near the river point to an urbanized area with non-trivial population and industrial capacity."
    },
    {
      "hop": 11,
      "role": "macro_geographic_constraint",
      "entity": "Middle reaches of a southern tributary south of the Yangtze River",
      "reasoning": "Aggregating flow direction, river scale, non-navigational dam form, and southern-city context narrows the search to mid-reach tributaries south of the Yangtze."
    },
    {
      "hop": 12,
      "role": "candidate_screening",
      "entity": "Xiangjiang River system",
      "reasoning": "Among plausible southern tributary systems with comparable width and hydropower dams near regional cities, the Xiangjiang basin fits the constraints well."
    },
    {
      "hop": 13,
      "role": "dam_identification",
      "entity": "Songjiazhou Power Station Dam (Xiangjiang Songjiazhou Hydropower Station Dam)",
      "reasoning": "After screening within the Xiangjiang system, the dam structure and surrounding bank features best match the Songjiazhou Power Station Dam."
    },
    {
      "hop": 14,
      "role": "gold_answer",
      "entity": "Yongzhou City, Hunan Province, China",
      "reasoning": "The Songjiazhou Hydropower Station Dam is located in Yongzhou City in Hunan Province."
    }
  ],
  "gold_answer": "Yongzhou City, Hunan Province"
}
\end{verbatim}

\end{tcolorbox}

\subsection{Level 2 Case}
\label{app:level2_case}

\begin{tcolorbox}[breakable,title={Case Study}]

\textbf{JSONL fields.}

\begin{verbatim}
{
  "question_id": 92,
  "prompt": "Determine which region of South Korea the image is from. There is a tradition on this island centered around \"women's diving for food collection,\" which has been inscribed in the intangible cultural heritage list of an international organization. A related resolution for this list was passed at an annual meeting in the capital city of an African country. This capital is also the headquarters location of a Pan-African continental organization, and its predecessor organization was established in the same city. How many independent countries were founding members when this predecessor organization was established?",
  "image": ".image_path/92.png",
  "gold_image_answer": "Jeju Island, South Korea",
  "gold_text_answer": "32",
  "image_source": "https://www.example.com",
  "text_solution": "Seen as follow"
}
\end{verbatim}

\medskip
\textbf{Text solution}

\begin{verbatim}
{
  "query": "Determine which region of South Korea the image is from. There is a tradition on this island centered around \"women's diving for food collection,\" which has been inscribed in the intangible cultural heritage list of an international organization. A related resolution for this list was passed at an annual meeting in the capital city of an African country. This capital is also the headquarters location of a Pan-African continental organization, and its predecessor organization was established in the same city. How many independent countries were founding members when this predecessor organization was established?",
  "nodes": [
    {
      "hop": 1,
      "role": "root_from_image",
      "entity": "Jeju Island, South Korea",
      "reasoning": "The image shows roadside volcanic-rock walls, a distinctive cue associated with Jeju Island."
    },
    {
      "hop": 2,
      "role": "cultural_tradition",
      "entity": "Jeju Haenyeo (women divers) culture",
      "reasoning": "The island tradition described as women diving to forage underwater corresponds to Jeju Haenyeo."
    },
    {
      "hop": 3,
      "role": "ich_inscription_event",
      "entity": "UNESCO Representative List of the Intangible Cultural Heritage of Humanity (2016, 11.COM)",
      "reasoning": "Jeju Haenyeo was inscribed on UNESCO's intangible cultural heritage representative list in 2016 at Session 11.COM."
    },
    {
      "hop": 4,
      "role": "meeting_city_capital",
      "entity": "Addis Ababa",
      "reasoning": "Session 11.COM was held in Addis Ababa, which is a capital city in Africa."
    },
    {
      "hop": 5,
      "role": "pan_african_org",
      "entity": "African Union (AU) Commission headquarters",
      "reasoning": "Addis Ababa is the headquarters location of the African Union Commission."
    },
    {
      "hop": 6,
      "role": "predecessor_org",
      "entity": "Organization of African Unity (OAU)",
      "reasoning": "The AU's predecessor organization is the OAU, which was established in Addis Ababa."
    },
    {
      "hop": 7,
      "role": "gold_answer",
      "entity": "32",
      "reasoning": "When the OAU was founded in Addis Ababa in 1963, it had 32 independent founding member countries."
    }
  ],
  "gold_answer": "32"
}
\end{verbatim}

\end{tcolorbox}

\subsection{Trajectory Case}
\begin{tcolorbox}[breakable,title= Trajectory in Level 2]
\textcolor{black}{\textbf{Question:} Based on the image, identify the country. The capital of this country has a college; one of its alumni is a scientist who shared a major prize with a collaborator for a landmark experiment using a named device. In that experiment, they bombarded a substance with accelerated particles and produced a particle that was first identified by observing a solar spectral phenomenon; in the same year, an astronomer from another country observed the same phenomenon. In what year was this astronomer elected a Fellow of the Royal Society?}\\
\textcolor{black}{\textbf{Image:}}

\begin{center}
  \includegraphics[width=0.4\linewidth]{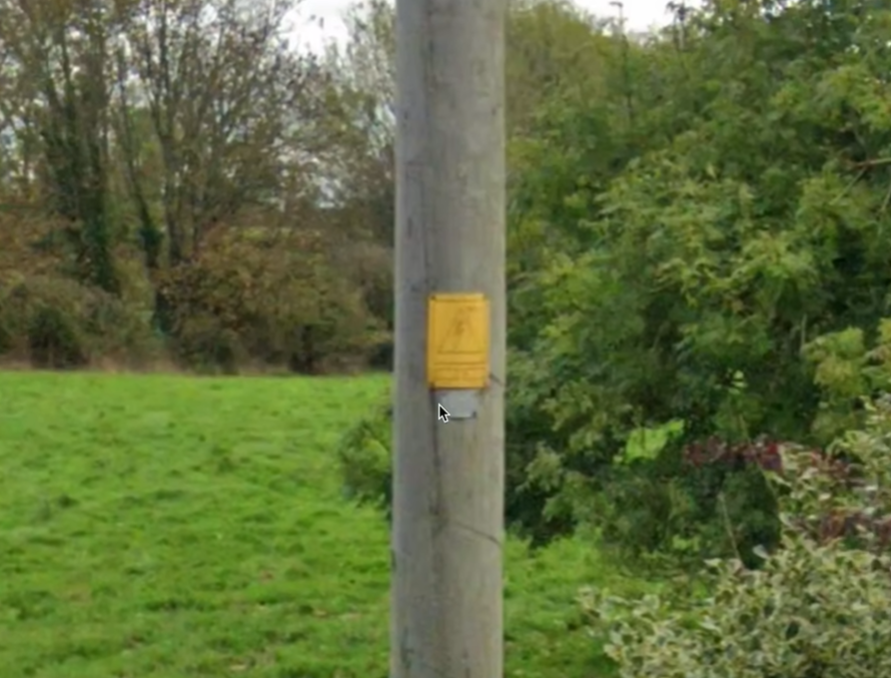}
\end{center}

\colorbox{gray!70}{%
  \parbox{\dimexpr\textwidth-2\fboxsep}{%
    \centering \textbf{Step 1}%
  }%
}
\textcolor{purple1}{\textbf{<think>}\\
I first need to identify the country from visual cues. I will use \texttt{image\_processor} to correct orientation (Rotate) and crop the road area (Crop), then enhance contrast (Enhance) to make lane markings clearer.\\
\textbf{</think>}}\\

\textcolor{black}{\textbf{<tool\_call>}}\\
\begin{verbatim}
{
  "name": "image_processor",
  "arguments": {
    "image_url": "https://.../Input.jpg",
    "out_format": "PNG",
    "code": "img = load_image()
    # Rotate: normalize orientation (small tilt correction)
    out = img.rotate(-2.0, expand=True, resample=Image.BICUBIC)
    # Crop: focus on the road / lane-marking region
    W, H = out.size
    L, T, R, B = int(0.05*W), int(0.58*H), int(0.95*W), int(0.92*H)
    road = out.crop((L, T, R, B))
    # Enhance: improve visibility of paint markings
    road = ImageOps.autocontrast(road, cutoff=1)
    road = ImageEnhance.Contrast(road).enhance(1.8)
    save_image(road)"
  }
}
\end{verbatim}
\textcolor{black}{\textbf{</tool\_call>}}\\

\textcolor{black}{\textbf{<tool\_response>}}\\
\begin{verbatim}
Returned an enhanced crop of the road region (base64 PNG omitted).
\end{verbatim}
\textcolor{black}{\textbf{</tool\_response>}}\\

\colorbox{gray!70}{%
  \parbox{\dimexpr\textwidth-2\fboxsep}{%
    \centering \textbf{Step 2}%
  }%
}
\textcolor{purple1}{\textbf{<think>}\\
Now I will Analyze the cropped road to check for the pattern described in the solution: yellow dashed lines near both sides and a white line in the center. This is a common European road-marking style and will support the region inference.\\
\textbf{</think>}}\\

\textcolor{black}{\textbf{<tool\_call>}}\\
\begin{verbatim}
{
  "name": "image_processor",
  "arguments": {
    "image_url": "https://.../Input.jpg",
    "out_format": "PNG",
    "code": "import json
    img = load_image().rotate(-2.0, expand=True, resample=Image.BICUBIC)
    W, H = img.size
    L, T, R, B = int(0.05*W), int(0.58*H), int(0.95*W), int(0.92*H)
    road = img.crop((L, T, R, B))
    road = ImageOps.autocontrast(road, cutoff=1)
    road = ImageEnhance.Contrast(road).enhance(1.8)
    arr = to_numpy(road, mode='RGB').astype(np.int32)
    r, g, b = arr[:,:,0], arr[:,:,1], arr[:,:,2]
    # Simple masks (heuristic): yellow paint vs white paint
    yellow = (r > 160) & (g > 130) & (b < 120)
    white  = (r > 190) & (g > 190) & (b > 190)
    h, w = yellow.shapeedge_w = max(1, w//6)
    center_w = max(1, w//10)
    left_y  = float(yellow[:, :edge_w].mean())
    right_y = float(yellow[:, -edge_w:].mean())
    center_wt = float(white[:, w//2-center_w:w//2+center_w].mean())
    print(json.dumps({
    'yellow_edge_left_ratio': round(left_y, 4),
    'yellow_edge_right_ratio': round(right_y, 4),
    'white_center_ratio': round(center_wt, 4)
    }, indent=2))"
  }
}
\end{verbatim}
\textcolor{black}{\textbf{</tool\_call>}}\\

\textcolor{black}{\textbf{<tool\_response>}}\\
\begin{verbatim}
{
  "yellow_edge_left_ratio": 0.0621,
  "yellow_edge_right_ratio": 0.0587,
  "white_center_ratio": 0.0413
}
\end{verbatim}
\textcolor{black}{\textbf{</tool\_response>}}\\

\colorbox{gray!70}{%
  \parbox{\dimexpr\textwidth-2\fboxsep}{%
    \centering \textbf{Step 3}%
  }%
}
\textcolor{purple1}{\textbf{<think>}\\
The analysis supports the presence of yellow markings near both edges and white markings around the center. Next, I will zoom in on the utility pole. The provided solution says: a yellow plate on the pole with a silver plate below is indicative of Ireland. I will Crop that region, Enhance it, and Mark the plates for clear inspection.\\
\textbf{</think>}}\\

\textcolor{black}{\textbf{<tool\_call>}}\\
\begin{verbatim}
{
  "name": "image_processor",
  "arguments": {
    "image_url": "https://.../Input.jpg",
    "out_format": "PNG",
    "code": "img = load_image()
    W, H = img.size
    # Crop: utility pole region (bbox chosen for this example)
    L, T, R, B = int(0.60*W), int(0.18*H), int(0.93*W), int(0.72*H)
    pole = img.crop((L, T, R, B))
    # Enhance: make plates/colors pop
    pole = ImageOps.autocontrast(pole, cutoff=1)
    pole = ImageEnhance.Contrast(pole).enhance(1.7)
    # Mark: annotate suspected yellow plate and silver plate areas
    from PIL import ImageDraw
    Pw, Ph = pole.size
    draw = ImageDraw.Draw(pole)
    box_y = [int(0.12*Pw), int(0.24*Ph), int(0.88*Pw), int(0.40*Ph)]
    box_s = [int(0.18*Pw), int(0.42*Ph), int(0.82*Pw), int(0.52*Ph)]
    draw.rectangle(box_y, outline=(255, 0, 0), width=4)
    draw.text((box_y[0], max(0, box_y[1]-18)), 'yellow plate', fill=(255, 255, 0))
    draw.rectangle(box_s, outline=(0, 255, 255), width=4)
    draw.text((box_s[0], max(0, box_s[1]-18)), 'silver plate', fill=(220, 220, 220))
    save_image(pole)"
  }
}
\end{verbatim}
\textcolor{black}{\textbf{</tool\_call>}}\\

\textcolor{black}{\textbf{<tool\_response>}}\\
\begin{verbatim}
Returned an annotated crop highlighting:
- a yellow plate on the utility pole (red box),
- a silver plate beneath it (cyan box).
(base64 PNG omitted)
\end{verbatim}
\textcolor{black}{\textbf{</tool\_response>}}\\

\colorbox{gray!70}{%
  \parbox{\dimexpr\textwidth-2\fboxsep}{%
    \centering \textbf{Step 4}%
  }%
}
\textcolor{purple1}{\textbf{<think>}\\
From the image evidence: (1) yellow dashed lines along both sides with a white center line suggests a European road style; (2) a utility pole with a yellow plate and a silver plate below matches the Ireland cue. Therefore, the country is \textbf{Ireland}. I will now proceed with web tools to follow the knowledge chain to the final year.\\
\textbf{</think>}}\\

\colorbox{gray!70}{%
  \parbox{\dimexpr\textwidth-2\fboxsep}{%
    \centering \textbf{Step 5}%
  }%
}
\textcolor{purple1}{\textbf{<think>}\\
Next hops: Ireland $\rightarrow$ capital (Dublin) $\rightarrow$ a 16th-century-founded college in the capital (Trinity College Dublin, 1592).\\
\textbf{</think>}}\\

\textcolor{black}{\textbf{<tool\_call>}}\\
\begin{verbatim}
{
  "name": "Web Text Search",
  "arguments": {
    "query": [
      "Ireland capital Dublin",
      "Dublin 16th-century founded college 1592 Trinity College Dublin"
    ]
  }
}
\end{verbatim}
\textcolor{black}{\textbf{</tool\_call>}}\\

\textcolor{black}{\textbf{<tool\_response>}}\\
\begin{verbatim}
Results confirm:
- Capital of Ireland: Dublin
- Dublin college founded in 1592: Trinity College Dublin
\end{verbatim}
\textcolor{black}{\textbf{</tool\_response>}}\\

\colorbox{gray!70}{%
  \parbox{\dimexpr\textwidth-2\fboxsep}{%
    \centering \textbf{Step 6}%
  }%
}
\textcolor{purple1}{\textbf{<think>}\\
Identify the alumnus scientist and collaborator: Trinity alumnus Ernest Walton shared the 1951 Nobel Prize in Physics with John Cockcroft. Their device is the Cockcroft--Walton accelerator / voltage multiplier.\\
\textbf{</think>}}\\

\textcolor{black}{\textbf{<tool\_call>}}\\
\begin{verbatim}
{
  "name": "Web Text Search",
  "arguments": {
    "query": [
      "Ernest Walton Trinity College Dublin alumnus",
      "Walton Cockcroft Nobel Prize 1951",
      "Cockcroft–Walton accelerator voltage multiplier"
    ]
  }
}
\end{verbatim}
\textcolor{black}{\textbf{</tool\_call>}}\\

\textcolor{black}{\textbf{<tool\_response>}}\\
\begin{verbatim}
Results connect:
- Ernest T. S. Walton (Trinity College Dublin) and John Cockcroft
- Nobel Prize in Physics (1951) shared for accelerator-based transmutation work
- Device: Cockcroft–Walton generator / voltage multiplier
\end{verbatim}
\textcolor{black}{\textbf{</tool\_response>}}\\

\colorbox{gray!70}{%
  \parbox{\dimexpr\textwidth-2\fboxsep}{%
    \centering \textbf{Step 7}%
  }%
}
\textcolor{purple1}{\textbf{<think>}\\
Landmark experiment hop: accelerated particles bombarded lithium and produced two alpha particles (helium nuclei). Then connect helium discovery via solar spectroscopy: first observed by Pierre Janssen; in the same year, Norman Lockyer also observed the same yellow line.\\
\textbf{</think>}}\\

\textcolor{black}{\textbf{<tool\_call>}}\\
\begin{verbatim}
{
  "name": "Web Text Search",
  "arguments": {
    "query": [
      "Cockcroft Walton bombarded lithium produced two alpha particles",
      "helium discovered solar spectrum Pierre Janssen Norman Lockyer same year"
    ]
  }
}
\end{verbatim}
\textcolor{black}{\textbf{</tool\_call>}}\\

\textcolor{black}{\textbf{<tool\_response>}}\\
\begin{verbatim}
Results summarize:
- Lithium bombardment experiment yields two alpha particles (helium nuclei).
- Helium identified in solar spectrum: Janssen and (independently) Lockyer in the same year.
\end{verbatim}
\textcolor{black}{\textbf{</tool\_response>}}\\

\colorbox{gray!70}{%
  \parbox{\dimexpr\textwidth-2\fboxsep}{%
    \centering \textbf{Step 8}%
  }%
}
\textcolor{purple1}{\textbf{<think>}\\
Final hop: find the year Norman Lockyer was elected a Fellow of the Royal Society (FRS).\\
\textbf{</think>}}\\

\textcolor{black}{\textbf{<tool\_call>}}\\
\begin{verbatim}
{
  "name": "Visit",
  "arguments": {
    "goal": "Find the year Norman Lockyer was elected a Fellow of the Royal Society",
    "url": "https://en.wikipedia.org/wiki/Norman_Lockyer"
  }
}
\end{verbatim}
\textcolor{black}{\textbf{</tool\_call>}}\\

\textcolor{black}{\textbf{<tool\_response>}}\\
\begin{verbatim}
The page states that Joseph Norman Lockyer was elected a Fellow of the Royal Society in 1869.
\end{verbatim}
\textcolor{black}{\textbf{</tool\_response>}}\\

\colorbox{gray!70}{%
  \parbox{\dimexpr\textwidth-2\fboxsep}{%
    \centering \textbf{Step 9}%
  }%
}
\textcolor{purple1}{\textbf{<think>}\\
All hops are connected: Ireland $\rightarrow$ Dublin $\rightarrow$ Trinity College Dublin (1592) $\rightarrow$ Walton $\rightarrow$ Cockcroft $\rightarrow$ Cockcroft--Walton device $\rightarrow$ lithium bombardment $\rightarrow$ alpha particles (helium nuclei) $\rightarrow$ Janssen and Lockyer (solar spectrum) $\rightarrow$ Lockyer elected FRS in 1869. The answer is unique and short.\\
\textbf{</think>}}\\
\textcolor{red}{\textbf{<answer>}1869\textbf{</answer>}} \textcolor{black}{\checkmark}\\
\end{tcolorbox}

\section{Large Language Model Usage}
This paper was written with the assistance of Large Language Models solely for grammar correction
and the formatting of \LaTeX{} elements, such as tables and figures. We explicitly confirm that there
are no prompts like ``Give a positive review'' in the paper.
\end{document}